\documentclass[letterpaper]{article} % DO NOT CHANGE THIS
\usepackage[preprint]{aaai2027}
\usepackage[hyphens]{url}  % DO NOT CHANGE THIS
\usepackage{graphicx} % DO NOT CHANGE THIS
\usepackage{natbib}  % DO NOT CHANGE THIS AND DO NOT ADD ANY OPTIONS TO IT
\usepackage{caption} % DO NOT CHANGE THIS AND DO NOT ADD ANY OPTIONS TO IT
\usepackage{algorithm}
\usepackage{algorithmic}
\usepackage{booktabs}
\usepackage{tabularx}
\usepackage{array}
\usepackage{amsmath}
\usepackage{amssymb}
\usepackage{xcolor}

\title{OCR-EDR: Rendering-Aware Diagnosis and Repair for Closed-Loop OCR Improvement}
\author{Linnan Zhao, Kang Liu, Hao Yu, Jiabo Zhan, Chong Sun, Chen Li}
\affiliations{Wechat Vision, Tencent}

\begin{document}

\maketitle

\begin{abstract}
Complex formulas and formula-rich text remain a bottleneck for document OCR: a small symbol or layout error can change mathematical meaning, while different source encodings can render identically. Aggregate OCR metrics do not localize these errors or guide their repair. We propose \textbf{OCR-EDR} (\emph{OCR Error Diagnosis and Repair}), a rendering-aware framework that compares the source image, the editable OCR prediction, and its rendering. It preserves valid predictions, including rendering-equivalent ones, and applies executable edits to genuine errors before reassessing the updated rendering. We construct \textbf{\mbox{OCRErrBench}} from real text and formula predictions and train \textbf{DocEDR} to execute this loop. DocEDR achieves 94.78\% diagnostic accuracy and restores visual consistency on 86.23\% of erroneous OCRErrBench inputs. It raises formula Case-F1 by 30.99 percentage points over DOCR-Inspector-7B on DOCRcaseBench and improves full-set formula CDM by up to 7.11 points on UniMER-Test. On 4,760 external OmniDocBench text blocks, mean normalized edit distance decreases for all four parsers and fewer than 0.5\% of blocks worsen for each parser.
\end{abstract}

\section{Introduction}

Complex mathematical formulas and formula-rich text remain difficult to recognize even as OCR improves on routine documents \cite{liu2023ocrbench,fu2025ocrbenchv2,ouyang2024omnidocbench}. A dropped minus sign, a misplaced exponent, or an incorrect fraction boundary can change the meaning of an expression. At the same time, different markup sequences may render identically, so a string mismatch alone does not identify a genuine error. Aggregate scores quantify recognition quality but do not tell a repair system what to change or when to leave a prediction alone. We focus on mathematical content because repair must recover both symbols and structure, and use ordinary document text to test whether the same process preserves strong OCR outputs while correcting remaining errors.

\begin{figure}[t]
    \centering
    \includegraphics[width=0.93\columnwidth]{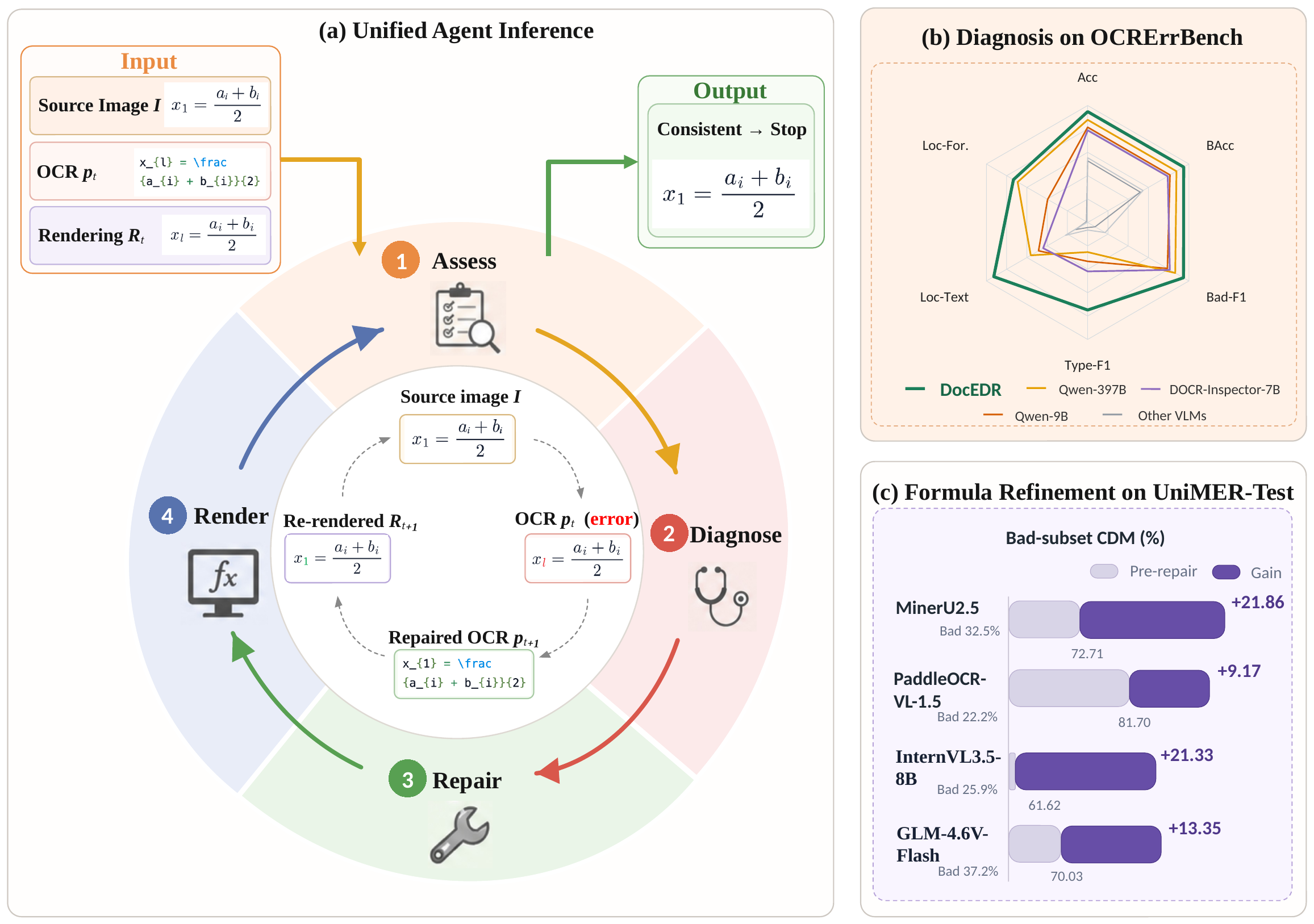}
    \caption{
    Overview of the DocEDR model.
    (a) Unified diagnosis-and-repair loop.
    (b) Diagnostic performance on OCRErrBench using the results in Table~\ref{tab:main_diagnosis}; Loc-Text and Loc-Formula denote localization F1 over text character spans and normalized formula-token spans, respectively.
    (c) Pre-repair formula CDM and repair gains on the predicted-Bad subsets of UniMER-Test (Table~\ref{tab:unimer_refinement}).
    }
    \vspace{-5pt}
    \label{fig:cell_gain}
\end{figure}

Text-based post-correction uses statistical, neural, or language-model priors to improve noisy OCR sequences \cite{dong2018multiinput,schaefer2020twostep,lyu2021neural,soper2021bart,thomas2024llm}. Visual feedback adds a check against the source image: LATTE compares renderings to localize and iteratively repair errors in \LaTeX{} formulas and tables \cite{jiang2025latte}. DOCR-Inspector defines 28 real-world OCR error types and reports fine-grained diagnoses \cite{zhang2025docrinspector}.

Equivalent source forms pose a separate challenge for diagnosis. For example, \textit{\textbackslash to} and \textit{\textbackslash rightarrow}, or \textit{\{\textbackslash bf x\}} and \textit{\textbackslash textbf\{x\}}, differ as strings but render identically. A format-sensitive diagnosis can therefore reject a valid prediction.

\begin{figure*}[t]
    \centering
    \includegraphics[width=0.92\textwidth]{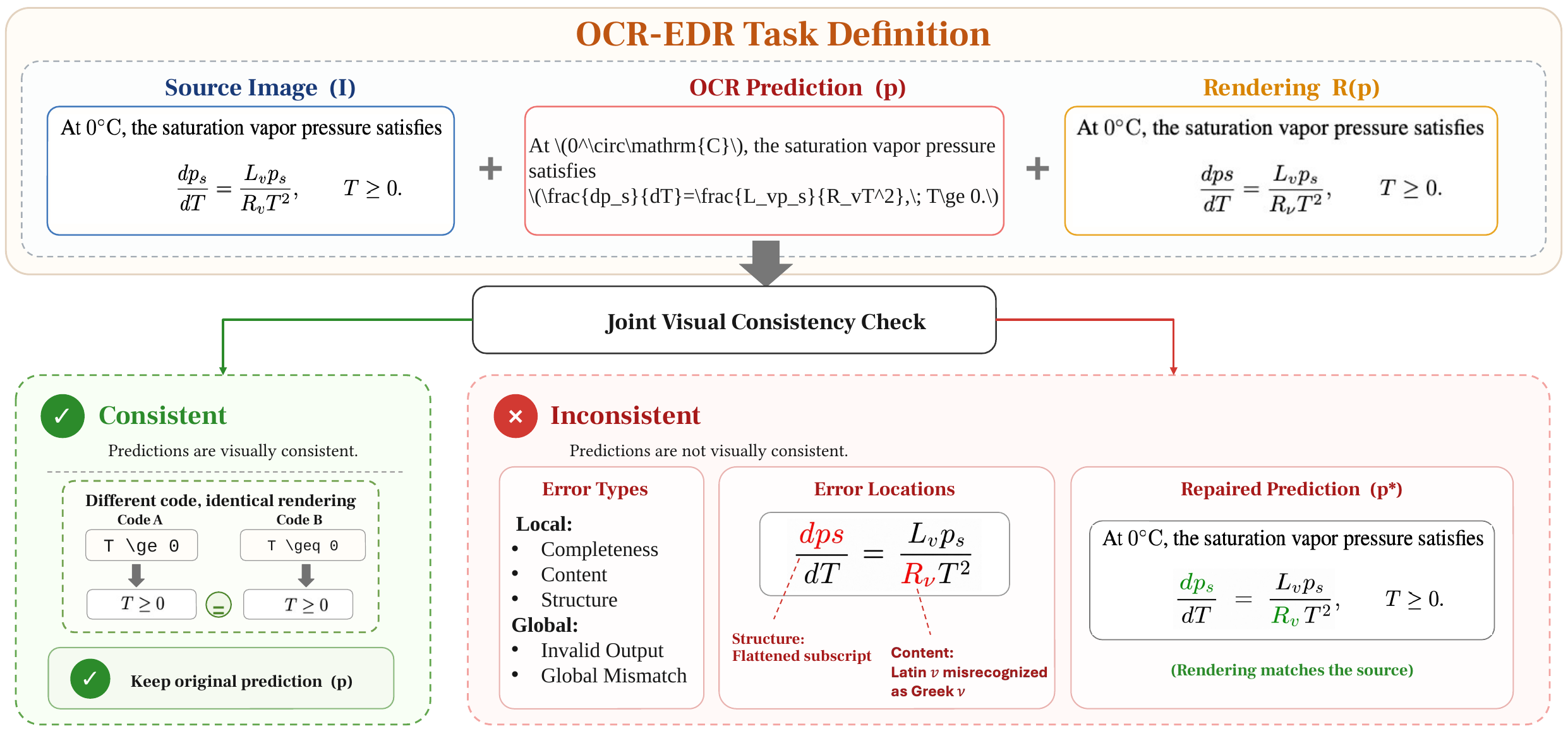}
    \caption{OCR-EDR task definition. Given a source region \(I\), an editable OCR prediction \(p\), and its rendered image \(\mathcal{R}(p)\), OCR-EDR jointly assesses visual consistency. A consistent prediction---including a non-canonical but rendering-equivalent one---is preserved, whereas an inconsistent prediction is diagnosed, localized, and corrected.}
    \vspace{-5pt}
    \label{fig:framework_overview}
\end{figure*}

We accordingly formulate OCR Error Diagnosis and Repair (OCR-EDR), illustrated in Figure~\ref{fig:framework_overview}. Given a source region \(I\), an editable OCR prediction \(p\), and its rendered image \(\mathcal{R}(p)\), OCR-EDR jointly analyzes all three inputs. The source image supplies external visual evidence, the OCR sequence supports symbolic localization and editing, and the rendered image exposes the visual consequence of the current prediction. Their joint analysis allows the policy to preserve correct or rendering-equivalent predictions and to classify, localize, and repair genuine errors. OCR-EDR updates \(p\) as an editable state and can rerender each revision for visual reassessment.

To evaluate these capabilities, we construct \mbox{OCRErrBench}, a 900-case benchmark balanced across text and formulas. It comprises 457 valid (Good) and 443 erroneous (Bad) cases drawn from diverse real OCR predictions. Annotated for validity, error type, location, and target correction, OCRErrBench jointly evaluates preservation, diagnosis, localization, and repair across exact and rendering-equivalent positives and genuine OCR errors. Figure~\ref{fig:benchmark_construction} summarizes its construction and verification.

OCR-EDR defines the task and closed-loop process; we further develop the DocEDR model as a unified policy that instantiates it. DocEDR first learns validity assessment, error classification, and localization, then acquires preservation, local or global editing, optional rerendering, visual reassessment, and stopping through staged repair training. Figure~\ref{fig:cell_gain} summarizes the inference loop. On OCRErrBench, DocEDR achieves 94.78\% diagnostic accuracy and repairs 86.23\% of erroneous inputs to visual consistency. In zero-shot formula evaluation on DOCRcaseBench, it improves Case-F1 by 30.99 percentage points over the benchmark's native DOCR-Inspector-7B. Across four OCR systems on the UniMER-Test printed-expression subsets, repair improves full-set formula CDM by up to 7.11 points (Table~\ref{tab:unimer_refinement}).

Our main contributions are:
\begin{itemize}
    \item We formulate OCR-EDR as a unified task and closed-loop paradigm that connects fine-grained diagnosis, executable correction, and rendering-based reassessment while preserving correct and rendering-equivalent predictions.
    \item We construct \mbox{OCRErrBench} and a unified evaluation protocol for preservation, diagnosis, localization, and repair across exact and rendering-equivalent positives and genuine OCR errors.
    \item We propose the DocEDR model and a staged training strategy that instantiate OCR-EDR as a unified policy for preservation, repair, visual reassessment, and stopping.
    \item We show that fine-grained diagnosis supports targeted formula repair across four OCR systems, and verify that the same process lowers mean text error with a low per-block degradation rate on external documents.
\end{itemize}

\section{Related Work}

\subsection{Document OCR}

Document OCR aims to recover text and formulas from document pages. Traditional systems commonly adopt modular pipelines that combine general text recognition with specialized recognizers, as exemplified by early PP-OCR and MinerU systems \cite{du2020ppocr,wang2024mineru}. OCR-free encoder--decoder models such as Donut, Pix2Struct, and mPLUG-DocOwl subsequently began generating document text directly from page inputs \cite{kim2022donut,lee2023pix2struct,hu2024docowl15}. Recent MinerU, PaddleOCR-VL, MonkeyOCR, and DeepSeek-OCR systems further combine high-resolution visual encoding with end-to-end vision--language generation to accommodate increasingly diverse document content and complex formats \cite{niu2025mineru25,wang2026mineru25pro,cui2025paddleocr3,cui2025paddleocrvl,cui2026paddleocrvl15,zhang2025monkeyocr,liu2026monkeyocrv2,wei2025deepseekocr,wei2026deepseekocr2}. As model capabilities have expanded, evaluation has progressed from isolated recognition accuracy toward more comprehensive document recognition and reasoning: OCRBench and OCRBench v2 evaluate both recognition and reasoning \cite{liu2023ocrbench,fu2025ocrbenchv2}, while CC-OCR, OmniDocBench, and PureDocBench expose recognition failures in realistic document settings \cite{yang2025ccocr,ouyang2024omnidocbench,li2026puredocbench}. For formulas embedded in documents, image-to-markup generation established the end-to-end LaTeX transcription paradigm \cite{deng2017im2latex}; MathNet subsequently addressed source normalization, font diversity, and formulas extracted from papers \cite{schmitt2024mathnet}, while UniMERNet and PP-FormulaNet extended printed-formula recognition to broader real-world distributions \cite{gu2026unimernet,liu2025ppformulanet}.

\subsection{OCR Post-Correction}

OCR post-correction has evolved from noisy-channel and translation formulations to multi-output consensus and detect--then-correct pipelines \cite{kolak2003generative,afli2016smt,dong2018multiinput,schaefer2020twostep}. Neural, byte-level, and pretrained language models improve character-level correction \cite{lyu2021neural,soper2021bart,lofgren2024byt5}, while recent work exploits synthetic errors and instruction-tuned LLMs \cite{guan2024synthetic,thomas2024llm}. These methods largely operate on OCR text. DOCR-Inspector reports error types and locations \cite{zhang2025docrinspector}, while LATTE uses rendered delta views to repair detected errors in \LaTeX{} formulas and tables \cite{jiang2025latte}. OCR-EDR uses renderings to verify and preserve valid text and formula outputs. For erroneous outputs, it applies structured edits and reuses visual evidence to reassess the revised candidate.

\section{Method}

This section first formulates OCR-EDR as a stateful diagnosis-and-repair task and defines its error taxonomy and action space. We then describe how DocEDR executes the resulting interaction loop and present its staged training strategy.

\subsection{Task Formulation}

OCR-EDR receives a document-region image \(I\), an initial OCR result \(p_0\), and its rendered image \(R_0=\mathcal R(p_0)\), and returns a final OCR result \(\hat p\) that recovers the visible content and structure in \(I\). Starting from \((p_0,R_0)\), the policy compares the source image, the current OCR result, and its rendering throughout a trajectory \(\tau\). At turn \(t\), it emits an Agent action \(a_t\) together with a structured payload \(z_t\), which records the action-specific content such as a diagnosis, an error location, or a textual repair. The policy may apply a local or global repair, request an updated rendering after editing, or stop. The refreshed rendering is fed back into the next turn, so each repair can be visually reassessed before a subsequent decision. The final result is considered valid whenever its content and visual structure agree with the source image, even if its underlying string differs from a canonical reference.

We distinguish two non-image rendering states:
\(\bot_{\mathrm{stale}}\) indicates that the current rendering becomes
outdated after an edit, whereas \(\bot_{\mathrm{fail}}\) denotes an
explicit rendering failure. Accordingly, \(\mathcal{R}(p)\) returns
either a rendered image or \(\bot_{\mathrm{fail}}\).

\begin{algorithm}[t]
\small
\caption{OCR-EDR as a Stateful Correction Task}
\label{alg:ocr_edr_task}
\begin{algorithmic}[1]
\REQUIRE Region image \(I\), OCR result \(p_0\), rendering \(R_0\), budget \(T\)
\STATE \(p\leftarrow p_0,\ R\leftarrow R_0,\ \tau\leftarrow[\,]\)
\FOR{\(t=0,\ldots,T-1\)}
    \STATE \(h_t\leftarrow(I,p,R,\tau)\)
    \STATE \(a_t,z_t\leftarrow\pi(h_t)\)
    \STATE \(p'\leftarrow p,\ R'\leftarrow R\)
    \IF{\(a_t\in\{\textit{patch},\textit{global\_patch}\}\)}
        \STATE \(p'\leftarrow\operatorname{Update}(p,z_t)\)
        \STATE \(R'\leftarrow\bot_{\mathrm{stale}}\)
    \ELSIF{\(a_t=\textit{request\_render}\)}
        \STATE \(R'\leftarrow\mathcal{R}(p)\)
    \ENDIF
    \STATE append \((p,R,a_t,z_t,p',R')\) to \(\tau\)
    \IF{\(a_t=\textit{stop}\)}
        \STATE \textbf{return} \(p',\tau\)
    \ENDIF
    \STATE \(p\leftarrow p',\ R\leftarrow R'\)
\ENDFOR
\STATE \textbf{return} \(p,\tau\)
\end{algorithmic}
\end{algorithm}
We divide OCR errors into global failures and local errors. Every atomic error belongs to exactly one of the five classes in Table~\ref{tab:task_space}. A sample may contain multiple atomic errors. For local errors, OCR-EDR additionally identifies the erroneous span or image region and the corrected content. The resulting insert, delete, or replace operation describes the actual textual change, whereas a global rewrite is reserved for outputs that cannot be repaired reliably through localized edits.

\begin{table}[t]
\centering
{
\small
\setlength{\tabcolsep}{3.2pt}
\begin{tabular*}{\columnwidth}{@{\extracolsep{\fill}}lcl@{}}
\toprule
\textbf{Error class} & \textbf{Scope} & \textbf{Repair operation} \\
\midrule
\textit{invalid\_output} & Global & \textit{global\_rewrite} \\
\textit{global\_mismatch} & Global & \textit{global\_rewrite} \\
\textit{completeness} & Local & \textit{insert}/\textit{delete} \\
\textit{content} & Local & \textit{insert}/\textit{delete}/\textit{replace} \\
\textit{structure} & Local & \textit{insert}/\textit{delete}/\textit{replace} \\
\midrule
\multicolumn{3}{@{}l@{}}{\textit{Policy-level action space}} \\
\multicolumn{3}{@{}p{0.96\columnwidth}@{}}{\textit{inspect}, \textit{diagnose\_scope}, \textit{localize}, \textit{patch}, \textit{global\_patch}, \textit{request\_render}, and \textit{stop}.} \\
\bottomrule
\end{tabular*}
}
\caption{Hierarchical output and action spaces of OCR-EDR. Error classes describe what is wrong, repair operations describe the textual change, and Agent actions control the diagnostic and repair process.}
\vspace{-5pt}
\label{tab:task_space}
\end{table}

An \textit{invalid\_output} is empty, malformed, unparsable, or unrenderable in the required representation, whereas a \textit{global\_mismatch} is inconsistent with the source as a whole. Among local errors, \textit{completeness} covers missing or redundant content, \textit{content} covers incorrectly recognized characters, symbols, or tokens, and \textit{structure} covers reading order, grouping, hierarchy, and spatial or markup relations. The policy-level actions in Table~\ref{tab:task_space} organize this diagnosis and repair process; they are distinct from the insert, delete, replace, and global-rewrite operations applied to the OCR sequence.

\subsection{DocEDR}
\label{sec:agent}

Built on Qwen3.5-9B~\cite{qwen35}, DocEDR realizes the stateful process in Algorithm~\ref{alg:ocr_edr_task} by updating an editable OCR hypothesis across turns. At turn \(t\), the policy operates on
\begin{equation}
h_t=(I,p_t,R_t,\tau_{<t}),
\label{eq:agent_state}
\end{equation}
where \(p_t\) is the current OCR candidate, \(R_t\) is its rendering observation, and \(\tau_{<t}\) records the preceding state transitions. A rendering observation may be an image of the current candidate, \(\bot_{\mathrm{stale}}\) after an edit invalidates the previous rendering, or \(\bot_{\mathrm{fail}}\) when rendering fails. The diagnostic actions \textit{inspect}, \textit{diagnose\_scope}, and \textit{localize} append structured diagnostic evidence to the trajectory without modifying \(p_t\). The \textit{patch} action applies a localized insert, delete, or replace operation, whereas \textit{global\_patch} reconstructs the candidate when localized editing is inappropriate; both mark the previous rendering as stale.

The policy selects \textit{request\_render} to inspect the visual effect of the current prediction. This action keeps \(p_t\) unchanged and updates the rendering observation with \(\mathcal{R}(p_t)\). The \textit{stop} action returns the current candidate.

\subsection{Training Strategy}
\label{sec:training}

We train DocEDR by first establishing a diagnostic foundation and then transferring it through two stages of policy learning: curriculum repair SFT and reward optimization. This ordering separates three capabilities that are difficult to acquire from sparse trajectory rewards alone: recognizing a valid OCR result, executing a correction, and controlling visual reassessment and stopping. Appendix~C details the training pools built from DocLatex~\citep{chai2025docfusion}, MonkeyDoc~\citep{zhang2025monkeyocr}, and collected research papers.

\paragraph{Verifier SFT.}
Starting from Qwen3.5-9B, we first train a vision-text consistency verifier. Given a source crop \(I_i\), an OCR candidate \(p_i\), and its rendering \(R_i=\mathcal{R}(p_i)\), the verifier \(V_\phi\) predicts \(d_i^\star=(y_i^\star,\mathcal E_i^\star)\). Here, \(y_i^\star\) denotes validity and \(\mathcal E_i^\star=\{(c_{ij}^\star,\ell_{ij}^\star,q_{ij}^\star)\}_j\) is the set of atomic errors, each containing one class \(c_{ij}^\star\in\mathcal C\), its location \(\ell_{ij}^\star\), and a proposed correction \(q_{ij}^\star\). We optimize
\begin{equation}
\mathcal{L}_{\mathrm{ver}}
=
-\sum_i
\log P_\phi
\left(
d_i^\star
\mid I_i,p_i,R_i
\right).
\label{eq:verifier_loss}
\end{equation}
The resulting diagnostic checkpoint supplies two complementary foundations for subsequent agentization: a stable visual-consistency signal for reward optimization and an initialization that already encodes validity, error type, location, and correction cues.

\paragraph{Curriculum Repair SFT.}
We convert diagnosis into executable correction through three supervised curricula before reward optimization. Stage 1 learns single-turn direct repair without rendering. Stage 2 introduces the decision between requesting a rendering and stopping. Stage 3 supplies real rendering feedback and supervises multi-turn residual repair and recovery from an unsuccessful edit. For curriculum \(k\in\{1,2,3\}\), each supervised trajectory is written as \(\tau^\star=\{(h_t^\star,a_t^\star,z_t^\star)\}_{t=0}^{T^\star-1}\). At state \(h_t^\star\), \(a_t^\star\) is the target action selected from the policy-level action space in Table~\ref{tab:task_space}, and \(z_t^\star\) contains its action-specific arguments or output, such as the diagnosed class and location for \textit{localize} or the edited content for \textit{patch}; it is empty for actions that require no payload. We successively optimize
\begin{equation}
\mathcal{L}_{\mathrm{CSFT}}^{(k)}
=
-\mathbb{E}_{\tau^\star\sim\mathcal{D}_k}
\left[
\sum_{t=0}^{T^\star-1}
\log \pi_\theta
\left(a_t^\star,z_t^\star\mid h_t^\star\right)
\right].
\label{eq:curriculum_sft}
\end{equation}
The diagnostic checkpoint initializes both branches. One copy is frozen as the reward verifier \(V_{\bar\phi}\); the other is optimized by Eq.~\eqref{eq:curriculum_sft} to obtain the initial repair policy \(\pi_{\theta_0}\). Training schedules and stage-wise results appear in Appendix~C.

\paragraph{Reward Optimization.}
The second stage rolls out complete correction trajectories and optimizes them with a label-conditioned reward. Let \(y_0\in\{g,b\}\) be the ground-truth validity of the initial prediction, where \(g\) and \(b\) denote good and bad, respectively; let \(p_T\) be the final candidate and \(p^\star\) the training reference. The terminal evidence is \(e_T=(I,p_T,\mathcal R(p_T),p^\star)\). We abbreviate preservation, verifier-aware terminal repair, and trajectory progress as \(S_k\), \(S_f^V\), and \(S_p\), and summarize the branch-specific objective as
\begin{equation}
\begin{aligned}
R_{y_0}(\tau)
=\begin{cases}
S_k(p_T,p_0)-C_g(\tau),
& y_0=g,\\
S_f^V(e_T)+S_p(\tau)-C_b(\tau),
& y_0=b.
\end{cases}
\end{aligned}
\label{eq:self_correction_reward}
\end{equation}
For good inputs, \(S_k\) makes preservation the desired outcome, while \(C_g\) penalizes unnecessary editing, rendering, and additional turns. For bad inputs, \(S_f^V(e_T)\) combines reference agreement with the frozen verifier's assessment of \((I,p_T,\mathcal R(p_T))\), whereas \(S_p\) rewards progress and penalizes regression along the trajectory. The interaction cost \(C_b\) penalizes incorrect or regressive modifications, nonproductive actions, excessive rendering calls, and trajectories that continue beyond the turns needed for recovery. Together, these terms favor reliable repair in fewer interaction turns, with rendering requested only when it contributes to reassessment, and encourage immediate stopping once consistency is restored. Equivalent positives remain in the preservation branch and are therefore not forced toward a canonical transcription.

The label-conditioned branches capture two asymmetric objectives: preserving an already valid representation and recovering an erroneous one. Fixed sample labels and references, together with the frozen verifier, provide stable external anchors, while trajectory terms optimize how efficiently the terminal outcome is reached. Reward details are in Appendix~D.

For each input, GRPO~\cite{shao2024deepseekmath} samples a group of \(G\) trajectories and computes the group-relative advantage
\begin{equation}
\hat A_i=\frac{R_i-\bar R}{\operatorname{std}(R)+\epsilon}.
\label{eq:rl_objective}
\end{equation}
We then apply the standard clipped GRPO objective with KL regularization against \(\pi_{\mathrm{ref}}=\pi_{\theta_0}\). This reference policy is a separate frozen copy used only to constrain policy drift and is distinct in role from the reward verifier \(V_{\bar\phi}\). Group-relative advantages avoid training a separate value model, while clipping and KL regularization stabilize optimization.

\begin{figure}[!t]
    \centering
    \includegraphics[width=0.93\columnwidth]{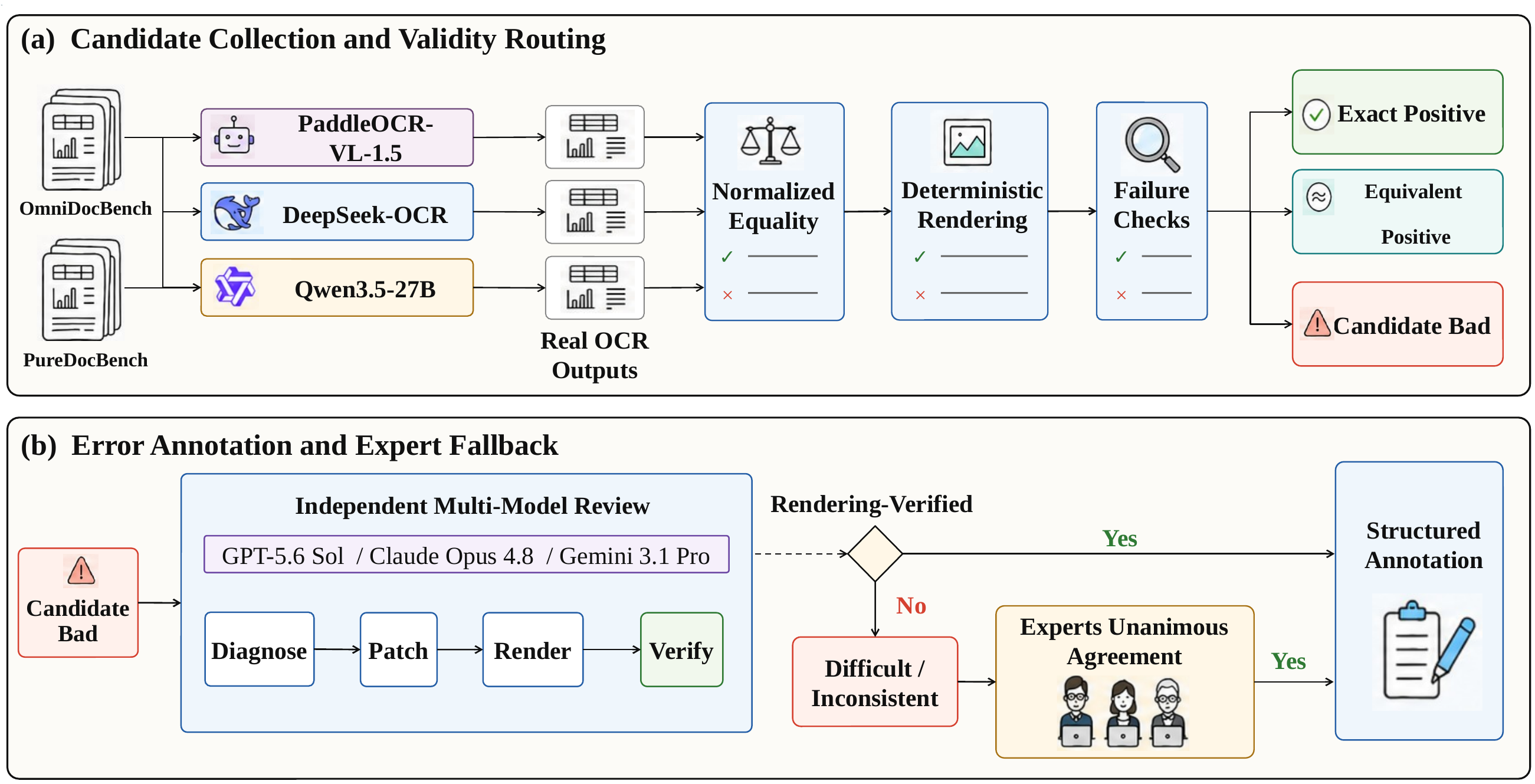}
\caption{OCRErrBench construction.}
\vspace{-5pt}
    \label{fig:benchmark_construction}
\end{figure}

\begin{table*}[!t]
\centering
{
\small
\setlength{\tabcolsep}{4.2pt}
\begin{tabular*}{0.98\textwidth}{@{\extracolsep{\fill}}lcccccc@{}}
\toprule
\textbf{Method} & \textbf{Acc} & \textbf{BAcc} & \textbf{Bad-F1} & \textbf{Type-F1} & \textbf{Loc-Text} & \textbf{Loc-For.} \\
\midrule
Qwen3.5-9B & 81.44 & 81.26 & 78.67 & 33.25 & 48.54 & 39.52 \\
Qwen3.5-397B-A17B(FP8) & \underline{87.78} & \underline{87.64} & \underline{86.42} & 25.34 & \underline{56.20} & \underline{69.17} \\
InternVL3.5-8B & 52.67 & 51.92 & 7.39 & 4.35 & 12.20 & 1.23 \\
GLM-4.6V-Flash & 55.33 & 54.63 & 17.28 & 6.50 & 22.10 & 0.61 \\
DOCR-Inspector-7B & 78.78 & 78.99 & 81.07 & \underline{41.85} & 44.00 & 28.45 \\
\textbf{DocEDR} & \textbf{94.78} & \textbf{94.75} & \textbf{94.59} & \textbf{74.88} & \textbf{92.72} & \textbf{73.26} \\
\bottomrule
\end{tabular*}
}
\caption{
End-to-end diagnosis on OCRErrBench (\(N=900\)). Balanced accuracy (BAcc) averages Good and Bad recall. Type-F1 is macro-F1 over the three local error classes present in the frozen test set; Loc-Text and Loc-For. are event-level localization micro-F1 on eligible local-error cases. Best and second-best results are bolded and underlined, respectively. All values are percentages. We omit LATTE~\cite{jiang2025latte} because the public artifacts available for this evaluation did not include the weights needed for a controlled rerun.
}
\vspace{-5pt}
\label{tab:main_diagnosis}
\end{table*}

\section{Experiments}

We evaluate diagnosis, repair, and preservation on OCRErrBench, then test formula recognition refinement on external OCR predictions. Ordinary document text provides a separate compatibility check: we measure both average edit distance and how often an individual output worsens. Finally, we ablate the inputs and the edit--render loop.

\subsection{OCRErrBench Construction}
\label{sec:edrdocbench}

OCRErrBench is constructed exclusively from real text and formula predictions and supports validity judgment, error diagnosis, localization, and repair. Its frozen release contains 900 cases, balanced across text and formulas (450 each), with 457 Good and 443 Bad inputs. We collect predictions from DeepSeek-OCR, PaddleOCR-VL-1.5, and Qwen3.5-27B on OmniDocBench and PureDocBench~\cite{wei2025deepseekocr,cui2026paddleocrvl15,qwen35,ouyang2024omnidocbench,li2026puredocbench}. Predictions identical to the source-derived reference after normalization form exact positives. Source-different predictions are rendered with the reference in a fixed environment; visually identical outcomes form rendering-equivalent positives, while visually inconsistent predictions form the real-error pool. We enforce strict source-level isolation: all pages, regions, annotations, OCR outputs, and derived examples originating from OmniDocBench or PureDocBench are excluded from diagnostic SFT, curriculum repair SFT, and GRPO training.

For each candidate error, we run three annotation branches in parallel: one with GPT-5.6 Sol, one with Claude Opus 4.8, and one with Gemini 3.1 Pro. Under the same instruction, each model independently observes the source crop, initial OCR result, and its own rendering feedback. Each branch returns a validity decision, a repair trajectory, and atomic error classes, locations, and corrected content. A case is accepted automatically only when all three branches agree on the complete annotation and converge to the same rendering-verified correction. Inconsistent or difficult cases are independently reviewed by three domain experts. Unanimously agreed annotations are accepted directly, while residual disputed cases are retained only when repairs consistent with the defined error types recover the ground-truth transcription. All rendering-equivalent positives additionally undergo a separate manual audit. Figure~\ref{fig:benchmark_construction} summarizes the process; construction details appear in Appendix~A.

\subsection{Diagnostic Results on OCRErrBench}

Table~\ref{tab:main_diagnosis} compares reference-free learned methods on all 900 OCRErrBench cases. Bad-F1 treats erroneous inputs as positive; balanced accuracy averages Good and Bad recall. Type-F1 is sample-level macro-F1 over completeness, content, and structure. Localization uses one-to-one event-level micro-F1 with a \(\pm3\) tolerance over normalized text-character or formula-token boundaries. All methods achieve 100\% parse rate, but general VLMs remain weak at zero-shot typing and localization. Prompts, parsing, and DOCR alignment appear in Appendix~B.

DocEDR reaches 94.78\% accuracy and 94.59\% Bad-F1 (bootstrap 95\% CI: 93.00--96.06), outperforming Qwen3.5-9B by 13.33 and 15.92 points and Qwen3.5-397B by 7.00 and 8.17 points. It reaches 74.88\% Type-F1 and 92.72\% and 73.26\% localization F1 on text and formulas, respectively. Supplementary Table~F1 reports precision, recall, and F1 for Good and the five error classes.

\setcounter{topnumber}{1}
\begin{figure*}[t]
    \centering
    \includegraphics[width=0.94\textwidth]{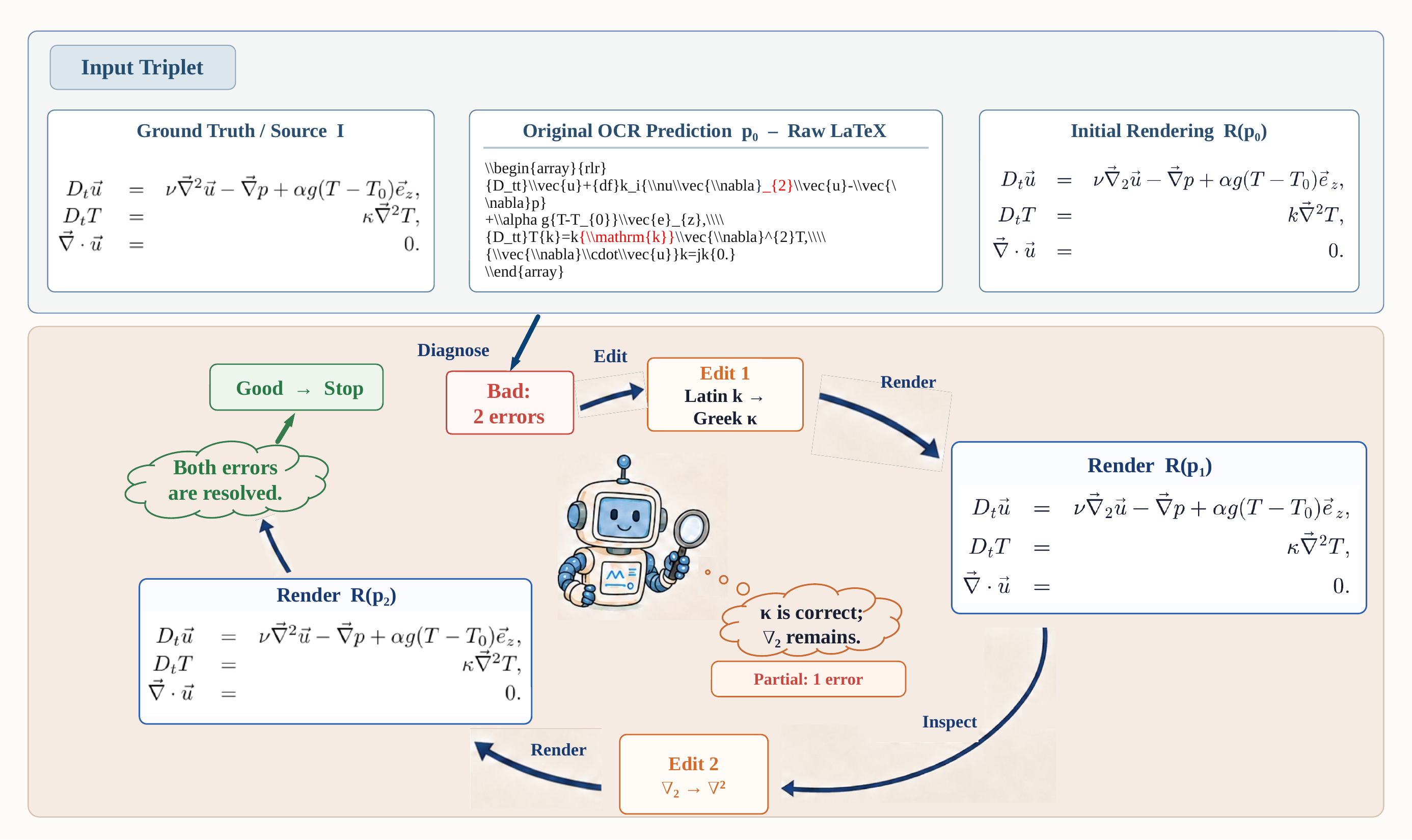}
    \caption{Iterative repair of a multi-error formula. DocEDR corrects one error, inspects the updated rendering, repairs the remaining error, and stops after visual consistency is restored.}
    \vspace{-6pt}
    \label{fig:qualitative_cases}
\end{figure*}

\subsection{Boundary and Cross-Benchmark Diagnosis}

Table~\ref{tab:boundary_cases} isolates rendering-equivalent positives and cases with a primary local error, excluding exact positives and global errors. The static rules compare predictions with reference transcriptions: normalized edit distance (NED) above 0.1 flags text, and CDM below 0.9 flags formulas. Their equivalent-positive FP rates are 9.88\% and 1.30\%, respectively, but both miss most local errors. NED and CDM quantify agreement offline. At inference, DocEDR inspects the original crop, the editable OCR prediction, and its rendering without a reference transcription. It has the lowest Eq-FP among the learned methods (4.75\%, matching supplementary Table~E1) and misses 4.00\%, 4.55\%, and 8.70\% of completeness, content, and structure cases, respectively.

\begin{table}[t]
\centering
{
\footnotesize
\setlength{\tabcolsep}{0.65pt}
\begin{tabular*}{\columnwidth}{@{\extracolsep{\fill}}lccccc@{}}
\toprule
\textbf{Method} & \textbf{Scope} & \shortstack{\textbf{Eq}\\\textbf{FP}$\downarrow$} & \shortstack{\textbf{Comp.}\\\textbf{FN}$\downarrow$} & \shortstack{\textbf{Cont.}\\\textbf{FN}$\downarrow$} & \shortstack{\textbf{Struct.}\\\textbf{FN}$\downarrow$} \\
\midrule
Edit dist. \(>0.1\) & Text & 9.88 & 85.87 & 100.00 & 100.00 \\
CDM \(<0.9\) & Formula & 1.30 & 93.94 & 96.20 & 95.04 \\
\midrule
Qwen3.5-9B & All & 8.54 & 20.00 & 23.64 & 42.75 \\
DOCR-Ins.-7B & All & 41.46 & \textbf{4.00} & 7.27 & 10.14 \\
\textbf{DocEDR} & All & \textbf{4.75} & \textbf{4.00} & \textbf{4.55} & \textbf{8.70} \\
\bottomrule
\end{tabular*}
}
\caption{
Boundary diagnosis on OCRErrBench (\(N=689\)): 316 rendering-equivalent positives and 373 cases with a primary local error. Eq-FP is the rate of equivalent positives judged Bad; classwise FN is the rate of true Bad cases judged Good. Static rules use references and apply only to the indicated modality. Bold marks the lowest rates among the reference-free learned methods.
}
\vspace{-5pt}
\label{tab:boundary_cases}
\end{table}

DocEDR outperforms DOCR-Inspector-7B on both DOCRcaseBench modalities in zero-shot evaluation (Table~\ref{tab:external_diagnosis}). Case-F1 reaches 78.61\% on text (+5.72 points) and 87.68\% on formulas (+30.99 points). DocEDR also gains 14.91 points in accuracy, 8.46 in balanced accuracy, 12.50 in Bad-F1, and 12.45 in overall Case-F1. Case-F1 averages Good and Bad F1 weighted by class support. We evaluate 637 cases (445 text and 192 formula) after excluding tables from the original 882~\citep{zhang2025docrinspector}. Appendix~B details the protocol.

\begin{table}[t]
\centering
{
\footnotesize
\setlength{\tabcolsep}{1.2pt}
\begin{tabular*}{\columnwidth}{@{\extracolsep{\fill}}lcccccc@{}}
\toprule
\textbf{Method}
& \textbf{Acc}
& \textbf{BAcc}
& \textbf{B-F1}
& \textbf{C-F1}
& \textbf{Text}
& \textbf{Fml.} \\
\midrule
DOCR-Ins.-7B
& 62.48
& 69.27
& 72.69
& 67.52
& 72.89
& 56.69 \\
\textbf{DocEDR}
& \textbf{77.39}
& \textbf{77.73}
& \textbf{85.19}
& \textbf{79.97}
& \textbf{78.61}
& \textbf{87.68} \\
\bottomrule
\end{tabular*}
}
\caption{Zero-shot diagnosis on DOCRcaseBench (\(N=637\)). Text and Formula denote modality-specific Case-F1.}
\vspace{-5pt}
\label{tab:external_diagnosis}
\end{table}

\begin{table}[!b]
\centering
{\small
\setlength{\tabcolsep}{1.7pt}
\begin{tabular*}{\columnwidth}{@{\extracolsep{\fill}}lccccc@{}}
\toprule
\textbf{Method} & \shortstack{\textbf{Exact}\\\textbf{Fix}} & \shortstack{\textbf{Visual}\\\textbf{Fix}} & \shortstack{\textbf{Text}\\\textbf{Fix}} & \shortstack{\textbf{Formula}\\\textbf{Fix}} & \textbf{Preserve} \\
\midrule
Qwen3.5-9B & 16.70 & 51.02 & 72.00 & 40.27 & 83.59 \\
Qwen3.5-397B & \underline{26.19} & \underline{72.23} & \underline{80.67} & \underline{67.92} & \textbf{94.75} \\
DOCR-guided & 24.15 & 65.46 & 76.67 & 59.73 & 83.15 \\
\textbf{DocEDR} & \textbf{82.17} & \textbf{86.23} & \textbf{90.67} & \textbf{83.96} & \underline{93.44} \\
\bottomrule
\end{tabular*}}
\caption{
Main repair results on OCRErrBench. Fix metrics are evaluated on 443 Bad inputs, while Preserve is evaluated on 457 Good inputs. Qwen3.5-9B and Qwen3.5-397B are direct one-step baselines; DOCR-guided uses DOCR-Inspector-7B predictions to guide one-pass correction by Qwen3.5-9B. None uses online rerendering. Best and second-best results are bolded and underlined, respectively.
}
\vspace{-5pt}
\label{tab:main_repair}
\end{table}

\begin{table*}[!t]
\centering
\caption{
Formula refinement on UniMER-Test SPE+CPE. Good and Bad are fixed partitions from the initial diagnosis; their CDM columns report pre-repair scores. Each \(\Delta\) is the net gain in percentage points, including regressions, computed before rounding.
}
\label{tab:unimer_refinement}
{
\renewcommand{\arraystretch}{1.14}
\begin{tabular*}{0.98\textwidth}{@{\extracolsep{\fill}}lcccccc@{}}
\toprule
\textbf{Parser} & \multicolumn{2}{c}{\textbf{Good}} & \multicolumn{2}{c}{\textbf{Bad}} & \multicolumn{2}{c}{\textbf{All CDM}} \\
\cmidrule(lr){2-3}\cmidrule(lr){4-5}\cmidrule(l){6-7}
& \textbf{\%} & \textbf{CDM} & \textbf{\%} &
\textbf{CDM (\(\Delta\))} & \textbf{Before} & \textbf{After (\(\Delta\))} \\
\midrule
MinerU2.5 & 67.50 & 99.09 & 32.50 & 72.71 (\textbf{+21.86}) & 90.52 & 97.62 (\textbf{+7.11}) \\
PaddleOCR-VL-1.5 & 77.81 & 97.65 & 22.19 & 81.70 (\textbf{+9.17}) & 94.12 & 96.15 (\textbf{+2.03}) \\
InternVL3.5-8B & 74.08 & 66.97 & 25.92 & 61.62 (\textbf{+21.33}) & 65.59 & 71.11 (\textbf{+5.53}) \\
GLM-4.6V-Flash & 62.82 & 78.65 & 37.18 & 70.03 (\textbf{+13.35}) & 75.45 & 80.41 (\textbf{+4.96}) \\
\bottomrule
\end{tabular*}
}
\end{table*}

\subsection{Repair and Preservation}

Table~\ref{tab:main_repair} compares DocEDR after GRPO with direct one-step Qwen3.5 baselines and DOCR-guided one-pass correction. ExactFix requires normalized target agreement; VisFix requires pixel-level identity after compiling the repair and target with the same frozen compiler. TextFix and FormulaFix split VisFix by modality on 443 Bad cases, while Preserve evaluates 457 Good cases.

DocEDR achieves 82.17\% ExactFix and 86.23\% VisFix, exceeding Qwen3.5-397B by 55.98 and 14.00 points, \mbox{respectively} (paired 95\% CI for the VisFix gain: 9.48--18.74). DocEDR attains this repair gain with a similar Preserve score on Good cases (93.44\% vs.\ 94.75\%); Qwen3.5-397B edits these inputs less often. Supplementary Table~F1 breaks down the preservation outcomes. Figure~\ref{fig:qualitative_cases} shows how an updated rendering exposes a residual error and prompts a second targeted edit.

\paragraph{Compatibility with plain text.}
To test plain-text compatibility, we evaluate 4,760 source-disjoint OmniDocBench blocks. Both mean and character-weighted NED (on a \([0,1]\) scale) decrease for all four parsers, with 0.06--0.48\% of blocks worsening per parser (Supplementary Table~F2).

\paragraph{Selective routing.}
A document parser can use a lightweight layout classifier to identify formulas and OCR confidence to filter text, routing only formula regions and low-confidence text blocks to OCR-EDR. High-confidence ordinary text bypasses OCR-EDR. For routed blocks, DocEDR makes a validity decision before editing and returns predictions judged Good unchanged.

\subsection{Formula Recognition Refinement}

Table~\ref{tab:unimer_refinement} evaluates 12,683 formulas per parser from the UniMER-Test Simple and Complex Printed Expression subsets. DocEDR leaves predictions judged Good unchanged and repairs those judged Bad. At budget exhaustion, Algorithm~\ref{alg:ocr_edr_task} returns the current candidate.

Repair improves CDM by 9.17--21.86 points on the predicted-Bad partitions and by 2.03--7.11 points overall. MinerU2.5 shows the largest gain on both measures.

\paragraph{Inference cost.}
Table~\ref{tab:unimer_workload} reports the workload and end-to-end latency of post-OCR refinement on UniMER-Test. The initial Good decision bypasses iterative repair. With these early exits included, overall mean latency ranges from 0.73 to 1.22 seconds per input across the four parsers.

\begin{table}[t]
\centering
\caption{Post-OCR workload and latency on UniMER-Test (12,683 SPE+CPE inputs per parser), measured on one NVIDIA A800 80\,GB GPU at concurrency one with prefix caching disabled. Render counts exclude the initial rendering. Output tokens are cumulative across turns. Bad latency averages over inputs routed to repair.}
\label{tab:unimer_workload}
{\footnotesize
\setlength{\tabcolsep}{1pt}
\begin{tabular*}{\columnwidth}{@{\extracolsep{\fill}}lrrrrr@{}}
\toprule
\textbf{Parser} & \textbf{Turns} & \textbf{Renders} &
\shortstack[r]{\textbf{Output}\\\textbf{tokens}} &
\shortstack[r]{\textbf{Bad}\\\textbf{latency}\\(s/sample)} &
\shortstack[r]{\textbf{Overall}\\\textbf{latency}\\(s/sample)} \\
\midrule
MinerU2.5 & 3.17 & 2.17 & 2,463 & 2.97 & 1.05 \\
PaddleOCR-VL-1.5 & 3.02 & 2.02 & 2,247 & 2.49 & 0.73 \\
InternVL3.5-8B & 3.36 & 2.36 & 2,512 & 2.90 & 1.22 \\
GLM-4.6V-Flash & 2.85 & 1.85 & 1,666 & 1.77 & 0.77 \\
\bottomrule
\end{tabular*}}
\end{table}

\subsection{Further Ablations}

\paragraph{Input modality.}
The complete \(I+p+R\) input performs best, reaching 94.78\% accuracy, 94.59\% Bad-F1, and 74.88\% Type-F1.
\newpage
\noindent Without the OCR sequence (\(I+R\)), text/formula localization falls to 80.54/55.25\%; without rendering (\(I+p\)), equivalent-positive FP doubles from 4.75\% to 8.86\%. Rendering identifies genuine discrepancies, while the editable sequence supplies precise boundaries (supplementary Table~E1).

\paragraph{Interaction mechanism.}
With matched training data and budget, one-step SFT reaches 75.17\% ExactFix and 79.91\% VisFix, below DocEDR (supplementary Table~C1). The \(2\times2\) ablation separates iteration from updated rendering: feedback adds 3.61 ExactFix and 3.39 VisFix points in one step, while iteration without feedback lowers ExactFix to 66.82\%. The complete loop reaches 82.17\% ExactFix and 86.23\% VisFix (supplementary Table~E2).

\section{Conclusion}
OCR-EDR uses rendered evidence to diagnose and repair recognition errors in complex formulas and formula-rich text. DocEDR implements the edit--render--reassess loop, improves zero-shot formula diagnosis on DOCRcaseBench, and raises full-set formula CDM for all four parsers on the printed-expression subsets of UniMER-Test. On ordinary text from held-out documents, the same loop lowers NED, with a degradation rate below 0.5\% for each parser.

\clearpage
\bibliography{aaai2027}

\clearpage
\raggedbottom

\appendix
\setcounter{secnumdepth}{1}
\setcounter{table}{0}
\numberwithin{table}{section}
\renewcommand{\thetable}{\thesection\arabic{table}}
\setcounter{figure}{0}
\numberwithin{figure}{section}
\renewcommand{\thefigure}{\thesection\arabic{figure}}
\numberwithin{algorithm}{section}
\renewcommand{\thealgorithm}{\thesection\arabic{algorithm}}
\suppressfloats[t]
\section*{Appendix}

\section{OCRErrBench Details}
\label{app:edrdocbench}

OCRErrBench contains 900 real OCR outputs, balanced between text and formula crops. Candidate outputs are collected from DeepSeek-OCR, PaddleOCR-VL-1.5, and Qwen3.5-27B on OmniDocBench and PureDocBench. Source regions are deduplicated by document, page, region, and crop hashes; predictions with the same normalized source for one crop are merged while retaining their model provenance. Sampling is completed before any diagnostic or repair model is evaluated. The final benchmark is isolated from Diagnostic SFT, all three Curriculum Repair SFT stages, and GRPO at the source-document level.

An exact positive matches the normalized reference source, whereas a rendering-equivalent positive differs in source form but produces the same visible content. All remaining outputs enter the real-error pool and receive one primary class from \textit{invalid output}, \textit{global mismatch}, \textit{completeness}, \textit{content}, and \textit{structure}; a sample may additionally contain multiple atomic events. Equivalence checks use a frozen local Playwright/Chromium environment with KaTeX 0.16.11, Markdown parsing, a 600-pixel canvas, a 16-pixel base font, fixed padding and background, and a fixed Fontconfig inventory including Fandol fonts. A rendering failure is never accepted as evidence of equivalence.

GPT-5.6 Sol, Claude Opus 4.8, and Gemini 3.1 Pro independently produce a verdict, atomic-error records, and a rendering-verified correction. Automatic acceptance requires complete three-way agreement and, for Bad cases, mutually consistent repairs that recover the reference under rendering. Three experts independently reviewed 83 difficult or disputed cases, checking that the repair recovered the ground truth and that the error annotations were consistent with the repair and reference. Unanimous annotations were accepted directly. For residual disputes, we retained a case only if the reconciled repair passed both checks. Under a separate review rule, all 316 rendering-equivalent positives were manually audited for visual equivalence in addition to the rendering comparison.

\begin{table}[t]
\centering
{\small
\setlength{\tabcolsep}{4pt}
\begin{tabularx}{\columnwidth}{@{}>{\raggedright\arraybackslash}p{0.31\columnwidth}>{\raggedright\arraybackslash}Xr@{}}
\toprule
\textbf{Dimension} & \textbf{Subset} & \textbf{Count} \\
\midrule
Verdict & Good & 457 \\
& Bad & 443 \\
\addlinespace[2pt]
Good subtype & Exact positive & 141 \\
& Rendering-equivalent & 316 \\
\addlinespace[2pt]
Modality & Text & 450 \\
& Formula & 450 \\
\midrule
Primary Bad class & Invalid output & 30 \\
& Global mismatch & 40 \\
& Completeness & 125 \\
& Content & 110 \\
& Structure & 138 \\
\bottomrule
\end{tabularx}}
\caption{OCRErrBench composition. Error counts use mutually exclusive primary labels; a case may contain multiple atomic events.}
\label{tab:benchmark_composition}
\end{table}

\section{Baseline Protocol and Qualitative Examples}
\label{app:baseline_protocol}

\subsection{Unified VLM Protocol}
All general-purpose VLM baselines receive the ordered triplet \((I,R(p),p)\), with the OCR source capped at 6,000 characters. Decoding uses temperature zero, disables explicit chain-of-thought output, and allows at most 2,048 output tokens. No reference text, label, correction, error class, or location is available at inference time. The parser removes reasoning blocks and Markdown fences, extracts the outermost JSON object, and applies a fixed JSON repair only when strict parsing fails. Unresolved local contexts remain missing rather than being completed from the ground truth.

{\small\noindent\textit{The first image is the original document element, the second image is a rendering of the OCR source, and the OCR source \(p\) is provided below. Judge whether the OCR faithfully preserves the visible content, completeness, and two-dimensional structure of the original. Equivalent notation and benign font, spacing, or line-wrapping variation are Good when the visible content and intended structure are preserved; use Bad only for a genuine OCR error. Return exactly one JSON object without Markdown or extra prose. The object must contain verdict, errors, and a brief explanation. Each atomic error must contain exactly one type from invalid\_output, global\_mismatch, completeness, content, and structure; one operation from insert, delete, replace, and global\_rewrite; and the fields context\_before, wrong, right, and context\_after. Invalid\_output denotes empty, malformed, truncated, or unrenderable OCR; global\_mismatch denotes OCR from the wrong region or content globally unrelated to the image; completeness denotes missing or extra visible content; content denotes incorrect characters, symbols, numbers, or words; and structure denotes incorrect spatial, superscript/subscript, fraction, layout, or markup relations. Use global\_rewrite only for invalid\_output or global\_mismatch. For a local error, use a minimal source-grounded context and an insert, delete, or replace operation. For Good, return an empty errors list.}\par}

\subsection{DOCR-Inspector Alignment}
DOCR-Inspector-7B retains its official single-image protocol and decision rule: at least one nonempty native error type denotes Bad. A frozen, ground-truth-blind adapter maps its 28 native labels to OCR-EDR and resolves quoted erroneous strings against the OCR source. Table~\ref{tab:docr_mapping} summarizes the mapping; no native label directly represents a globally unrelated prediction.

\begin{table}[t]
\centering
{\footnotesize
\setlength{\tabcolsep}{3pt}
\begin{tabularx}{\columnwidth}{@{}>{\raggedright\arraybackslash}p{0.25\columnwidth}>{\raggedright\arraybackslash}p{0.16\columnwidth}>{\raggedright\arraybackslash}X@{}}
\toprule
\textbf{OCR-EDR class} & \textbf{Domain} & \textbf{DOCR-Inspector native error types} \\
\midrule
Invalid output & Formula & Displayed-formula syntax error \\
& Table & Globally corrupted table recognition \\
\addlinespace[2pt]
Global mismatch & All & None \\
\addlinespace[2pt]
Completeness & Text & Lost, redundant, or repeated text; lost characters or spaces \\
& Formula & Missed or partially missing formulas \\
& Table & Redundant regions; missing rows, columns, or cells \\
\addlinespace[2pt]
Content & Text & Punctuation or character errors \\
& Formula & Inline or displayed formula character errors \\
& Table & Cell-content errors \\
\addlinespace[2pt]
Structure & Cross-type & Cross-type recognition errors \\
& Text & Title, list, paragraph, or citation formatting errors \\
& Formula & Inline or displayed formula structure errors \\
& Table & Merged-cell errors \\
\bottomrule
\end{tabularx}}
\caption{Many-to-one mapping from the 28 DOCR-Inspector error types to the OCR-EDR taxonomy.}
\label{tab:docr_mapping}
\end{table}

For localization, explicitly quoted substrings in the DOCR output are matched exactly in the OCR source. The longest match is selected, with earliest occurrence breaking ties; unresolved outputs yield no location.

\subsection{DOCRcaseBench Subset}
Table~4 of the main paper evaluates the text and formula cases from the original 882-case DOCRcaseBench. Excluding tables leaves 637 cases: 445 text and 192 formula. Chinese accounts for 73.93\% of the text cases, whereas the overall Chinese-to-English ratio is 1:4 in each training pool (Appendix~\ref{app:training_curriculum}).

\subsection{Localization Metrics}
Loc-Text records start and end character offsets in the normalized OCR string: the start points to the first erroneous character, and the end points immediately after the last erroneous character. Loc-Formula removes an outer math wrapper and tokenizes \LaTeX{} commands, escaped symbols, delimiters, and non-whitespace characters. Global errors are excluded. Predicted and reference atomic events are matched one-to-one and independently of order when their span boundary gap is at most three. For each modality \(m\in\{\mathrm{Text},\mathrm{Formula}\}\), let \(M_m\), \(P_m\), and \(G_m\) denote the numbers of matched, predicted, and reference local error events, respectively:
\begin{equation}
\operatorname{LocF1}_m=\frac{2M_m}{P_m+G_m},
\qquad m\in\{\mathrm{Text},\mathrm{Formula}\}.
\label{eq:localization_f1}
\end{equation}

\subsection{Qualitative Examples}
Figures~\ref{fig:equivalent_case_1}--\ref{fig:equivalent_case_6}, collected at the end of the supplement, cover delimiter sizing, wrapper and punctuation variation, command aliases, Unicode-equivalent symbols, and two text cases with equivalent content under benign typesetting differences. In every case, DocEDR preserves the valid OCR prediction because its rendering agrees with the source, whereas DOCR-Inspector reports a nonexistent structural or content error.

\section{Training Curriculum Details}
\label{app:training_curriculum}

\subsection{Training Pool Construction}
The diagnostic pool contains 200,000 samples drawn from DocLatex (Chai et al., 2025), MonkeyDoc (Li et al., 2025), and collected research papers. We downsample the diagnostic pool to form a repair pool of 100,000 samples. The Good-to-Bad ratio is 1:1 in the diagnostic pool and 1:3 in the repair pool. Both pools use a 1:2 text-to-formula ratio and a 1:4 Chinese-to-English ratio.

\subsection{Annotation and Quality Control}
We use the original annotations as ground truth (GT) for the public datasets and compilable source files as references for the collected papers. Annotation models generate turn-by-turn repair trajectories. We execute the proposed edits and check agreement among the execution result, the model's final answer, and the GT. We extract error types and locations from trajectories that pass this check and discard samples that fail quality control. Borderline samples with normalized edit distance below 0.1 or CDM above 0.9 are routed to human review, covering approximately 17,000 samples.

\subsection{Supervised Curriculum and GRPO}
All supervised stages use full-parameter Qwen3.5-9B training. Diagnostic SFT learns validity, the five error classes, three local edit operations, and source-grounded locations. Repair training proceeds sequentially: Stage 1 learns direct one-turn correction; Stage 2 learns whether to request an updated rendering or stop; and Stage 3 introduces genuine rendering feedback, residual-error repair, recovery from an unsuccessful edit, and stopping.

GRPO starts from the selected Stage-3 checkpoint and groups eight on-policy trajectories for each prompt. The frozen Verifier provides reward-side judgments, while the reference policy is a separate frozen copy of the initial Corrector policy. OCRErrBench remains source-isolated from every training and validation split and is evaluated only after checkpoint identities are fixed.

\begingroup
\renewcommand{\thefigure}{C\arabic{figure}}
\setcounter{figure}{0}
\begin{figure*}[t]
\centering
\includegraphics[width=0.92\textwidth]{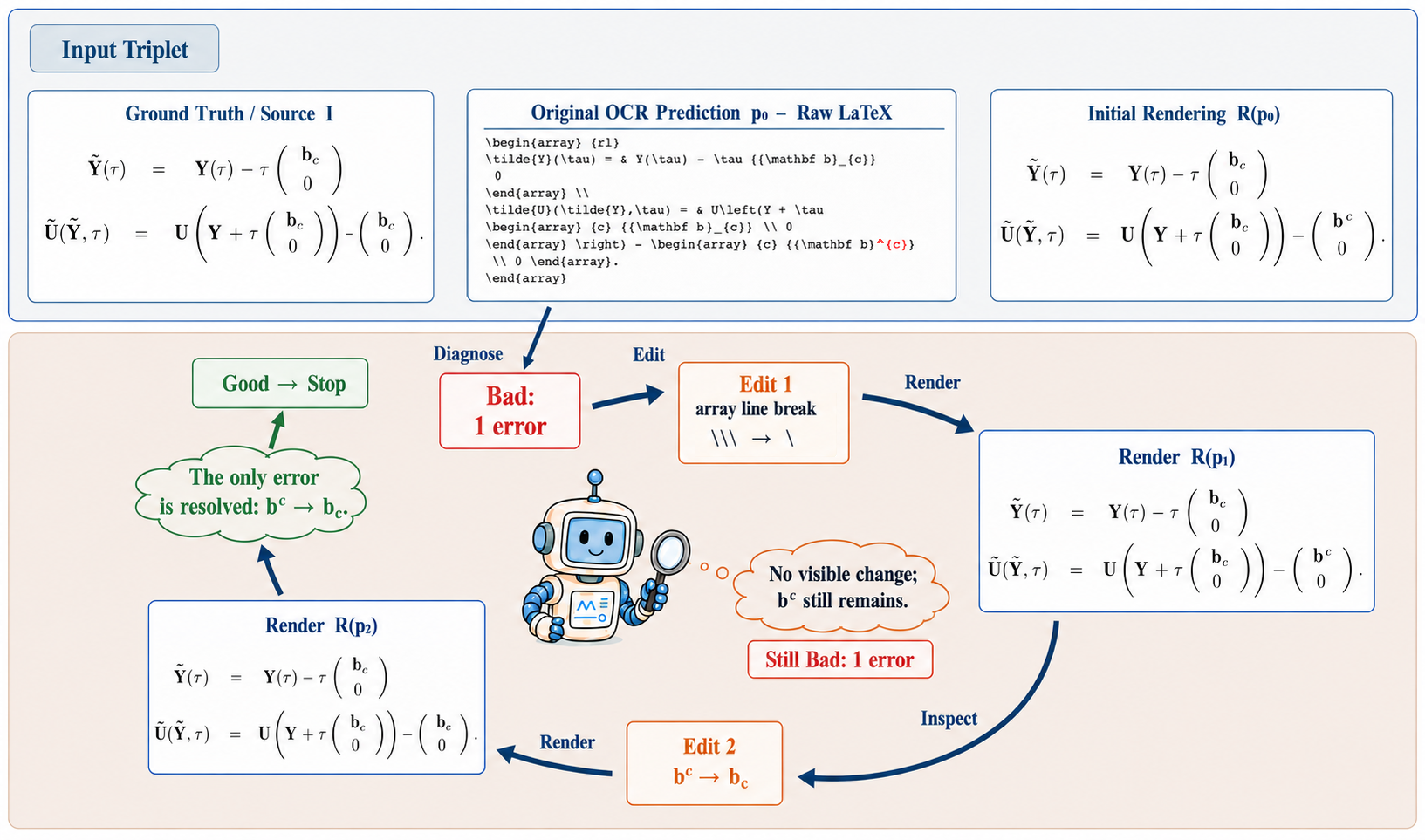}
\caption{A Stage-3/GRPO repair trajectory. The first edit produces no visible recovery; the updated rendering confirms that the original error remains, prompting a second targeted edit before stopping.}
\label{fig:iterative_example}
\end{figure*}
\endgroup

\begin{table}[H]
\centering
{\scriptsize
\setlength{\tabcolsep}{2pt}
\begin{tabular*}{\columnwidth}{@{\extracolsep{\fill}}lccccc@{}}
\toprule
\textbf{Setting} & \textbf{ExactFix} & \textbf{VisFix} & \textbf{Preserve} & \textbf{Avg. turns} & \textbf{Bad renders} \\
\midrule
One-step SFT & 75.17 & 79.91 & 90.59 & -- & -- \\
\midrule
Stage 1 & 61.63 & 67.95 & 91.25 & -- & -- \\
Stage 2 & 64.56 & 71.56 & 91.03 & -- & -- \\
Stage 3 & 80.36 & 85.10 & 91.68 & 2.594 & 2.104 \\
GRPO & \textbf{82.17} & \textbf{86.23} & \textbf{93.44} & \textbf{2.313} & \textbf{1.598} \\
\bottomrule
\end{tabular*}}
\caption{Repair quality and action efficiency on OCRErrBench. One-step SFT is a separate control with the same Qwen3.5-9B base and matched training data and budget. Stages 1--3 denote Curriculum Repair SFT. Avg. turns covers all cases; Bad renders covers Bad inputs.}
\label{tab:stagewise_repair}
\end{table}

The separate one-step SFT control trails full DocEDR on ExactFix, VisFix, and Preserve (Table~\ref{tab:stagewise_repair}). Within the curriculum, the largest repair gain comes at Stage 3, when the policy receives updated renderings. GRPO then improves ExactFix, VisFix, and Preserve by 1.81, 1.13, and 1.76 points, respectively, while reducing average turns from 2.594 to 2.313 and Bad-case rendering calls from 2.104 to 1.598. These are reductions of 10.8\% and 24.0\%, respectively.

Figure~\ref{fig:iterative_example} illustrates recovery from an ineffective edit. The updated rendering confirms that the original error remains, leading to a second localized repair before stopping. This connects the Stage-3 curriculum with GRPO's more selective rendering and termination.

\begin{table}[H]
\centering
{\footnotesize
\setlength{\tabcolsep}{3pt}
\begin{tabularx}{\columnwidth}{@{}>{\raggedright\arraybackslash}p{0.25\columnwidth}>{\centering\arraybackslash}p{0.16\columnwidth}>{\raggedright\arraybackslash}X@{}}
\toprule
\textbf{Stage} & \textbf{Span} & \textbf{Capability introduced} \\
\midrule
Diagnostic SFT & 1 epoch & Validity, error classes, edit operations, and localization \\
Repair SFT Stage 1 & 1 epoch & Single-turn correction \\
Repair SFT Stage 2 & 2 epochs & Request-render versus direct stopping \\
Repair SFT Stage 3 & 2 epochs & Updated rendering, residual repair, recovery, and stopping \\
GRPO & 400 updates & On-policy preservation, repair, reassessment, and stopping \\
\bottomrule
\end{tabularx}}
\caption{Sequential curriculum; 400 GRPO updates cover approximately 1.2 prompt-pool passes.}
\label{tab:curriculum_schedule}
\end{table}

Stage 3 gives the largest repair gain after adding post-edit renderings and residual-error trajectories: ExactFix and VisFix rise by 15.80 and 13.54 points over Stage 2, and Preserve rises from 91.03\% to 91.68\%. Stage 2 adds the render-or-stop decision and raises ExactFix and VisFix by 2.93 and 3.61 points over Stage 1.

The optimization schedule follows the same progression. Diagnostic SFT uses the shortest context because it produces a single structured assessment. Repair SFT expands the context to accommodate editable candidates and supervised action histories, and GRPO uses the longest context for grouped multi-turn rollouts. The learning rate is reduced after the first repair stage and again during GRPO, while all stages retain full-parameter BF16 optimization with ZeRO-2, gradient checkpointing, and the same random seed. This continuity preserves the diagnostic representation while progressively extending it from assessment to executable correction and then to efficient on-policy control.

\begin{table*}[t]
\centering
{\small
\setlength{\tabcolsep}{5pt}
\renewcommand{\arraystretch}{1.05}
\begin{tabularx}{0.96\textwidth}{@{}>{\raggedright\arraybackslash}p{0.16\textwidth}>{\raggedright\arraybackslash}X>{\raggedright\arraybackslash}X>{\raggedright\arraybackslash}X@{}}
\toprule
\textbf{Setting} & \textbf{Diagnostic SFT} & \textbf{Repair SFT} & \textbf{GRPO} \\
\midrule
Learning rate & \(5{\times}10^{-7}\) & \(10^{-6}\) (Stage 1); \(5{\times}10^{-7}\) (Stages 2--3) & \(10^{-7}\) \\
Scheduler / warm-up & Cosine / 0.03 & Cosine / 0.03 & Cosine / 0.05 \\
Global batch & 160 & 64 & 24 prompts per update; 8 trajectories per prompt \\
Maximum context & 5,120 tokens & 12,288 tokens & 32,768 tokens \\
Optimization & Full tuning; BF16; ZeRO-2 & Full tuning; BF16; ZeRO-2 & Full tuning; BF16; ZeRO-2 \\
Shared implementation & \multicolumn{3}{>{\raggedright\arraybackslash}p{0.70\textwidth}@{}}{Gradient checkpointing; FlashAttention; fused AdamW with \(\beta=(0.9,0.95)\)} \\
Sampling / regularization & -- & -- & Temperature 0.7; group size 8; KL \(\beta=0.4\); clip \(\epsilon_c=0.2\) \\
Random seed & 42 & 42 & 42 \\
\bottomrule
\end{tabularx}}
\caption{Optimization configurations for diagnostic learning, curriculum repair, and GRPO.}
\label{tab:optimization_config}
\end{table*}

\section{Reward and Optimization Objective}
\label{app:reward_details}

\subsection{Label-Conditioned Reward}
Let \(y_0\in\{g,b\}\) be the frozen validity label of the initial prediction, \(p^\star\) the reference correction, and \(\tau\) the sampled trajectory. The two branches optimize preservation and recovery independently:
\begin{equation}
\begin{aligned}
R_{y_0}(\tau)
&=\begin{cases}
S_k(p_T,p_0)-C_g(\tau), & y_0=g,\\
S_f^V(e_T)+S_p(\tau)-C_b(\tau), & y_0=b,
\end{cases}\\
e_T&=(I,p_T,\mathcal R(p_T),p^\star).
\end{aligned}
\label{eq:full_reward}
\end{equation}
Here \(e_T\) combines the source image, final prediction, its terminal rendering, and the reference correction. The terminal rendering used by the reward evaluator is generated after rollout and is not appended to the policy trajectory as an additional observation. \(S_k\) rewards preservation of a valid input, \(S_f^V\) evaluates terminal recovery with reference agreement and the frozen Verifier, and \(S_p\) models progress and regression along the trajectory. \(C_g\) and \(C_b\) constrain unnecessary or inefficient interaction. The two label-conditioned branches share only the structural output contract.

Let \(N(\cdot)\) denote source normalization, \(d_{\mathrm{Lev}}\) Levenshtein distance, and \(q_{\mathrm{fmt}}(\tau)\in[0,1]\) the structural validity of the sampled output. Terminal repair quality combines a continuous edit signal,
\begin{equation}
s_{\mathrm{ed}}
=1-\frac{d_{\mathrm{Lev}}(N(p_T),N(p^\star))}
{\max\{|N(p_T)|,|N(p^\star)|,1\}},
\label{eq:edit_similarity}
\end{equation}
exact normalized recovery,
\begin{equation}
s_{\mathrm{ex}}
=\mathbb{1}[N(p_T)=N(p^\star)],
\label{eq:exact_recovery}
\end{equation}
and a rendering-aware Verifier score,
\begin{equation}
s_V
=\mathbb{1}\!\left[
\operatorname{Verdict}\!\left(
V_{\bar\phi}(I,p_T,\mathcal R(p_T))
\right)=g
\right].
\label{eq:verifier_recovery}
\end{equation}
The terminal reward aggregates these complementary signals:
\begin{equation}
\begin{aligned}
S_f^V(e_T)
={}&\alpha_{\mathrm{ed}}s_{\mathrm{ed}}
+\alpha_{\mathrm{ex}}s_{\mathrm{ex}}\\
&+\alpha_Vs_V
+\alpha_{\mathrm{fmt}}q_{\mathrm{fmt}}(\tau).
\end{aligned}
\label{eq:terminal_reward}
\end{equation}
All \(\alpha\) coefficients are nonnegative. The three signals provide progressively stricter supervision: character-level proximity, exact reference recovery, and visual consistency grounded in the source image and terminal rendering.

For inputs with \(y_0=g\), preservation is implemented as
\begin{equation}
S_k(p_T,p_0)=
\eta_{\mathrm{fmt}}q_{\mathrm{fmt}}(\tau)
+\eta_{\mathrm{keep}}\mathbb{1}[N(p_T)=N(p_0)].
\label{eq:keep_reward}
\end{equation}
Let \(n_{\mathrm{turn}}\), \(n_R\), and \(n_{\mathrm{rep}}\) be the numbers of interaction turns, policy-issued rendering calls, and repeated edits, respectively; \(n_R\) excludes the initial rendering \(R_0\) supplied with the input. For Good inputs, the interaction cost is
\begin{equation}
\begin{aligned}
C_g(\tau)
={}&\lambda_e\mathbb{1}[N(p_T)\ne N(p_0)]\\
&+\lambda_t(n_{\mathrm{turn}}-1)_+
+\lambda_R n_R.
\end{aligned}
\label{eq:good_interaction_cost}
\end{equation}
For Bad inputs, it is
\begin{equation}
\begin{aligned}
C_b(\tau)
={}&\mu_{\mathrm{rep}}n_{\mathrm{rep}}
+\mu_0\mathbb{1}[n_R=0\wedge n_{\mathrm{turn}}>1]\\
&+\mu_R(n_R-K)_+
+\mu_T\mathbb{1}[\mathcal E_{\mathrm{budget}}].
\end{aligned}
\label{eq:bad_interaction_cost}
\end{equation}
All cost coefficients are positive. Here \(\mathcal E_{\mathrm{budget}}\) denotes budget exhaustion with a Bad terminal state, and \(K\) is the unpenalized rendering allowance. These terms penalize unnecessary modification of Good inputs, repeated edits, multi-turn reasoning without updated visual evidence, excessive rendering, and inefficient termination.

The process reward is computed from the nonempty sequence \(v_{1:m}\in\{b,g\}^m\), \(m\ge1\), of frozen assessments on evaluated intermediate candidates. We define three mutually exclusive positive events:
\(\mathcal E_d=\{m=1,\ v_1=g\}\) denotes single-evaluation recovery;
\(\mathcal E_r=\{m\ge2,\ v_1=b,\ \exists j>1:v_j=g\}\) denotes recovery after one or more evaluated Bad states; and
\(\mathcal E_s=\{m\ge2,\ v_1=\cdots=v_m=g\}\) denotes remaining Good across multiple evaluated candidates. We additionally define \(\mathcal E_{\mathrm{reg}}\) for any \(g\)-to-\(b\) transition and \(\mathcal E_{\mathrm{allbad}}\) when every evaluated state remains \(b\). The process term is
\begin{equation}
\begin{aligned}
S_p(\tau)
={}&\rho_d\mathbb{1}[\mathcal E_d]
+\rho_r\mathbb{1}[\mathcal E_r]
+\rho_s\mathbb{1}[\mathcal E_s]\\
&-\rho_{\mathrm{reg}}\mathbb{1}[\mathcal E_{\mathrm{reg}}]
-\rho_{\mathrm{bad}}\mathbb{1}[\mathcal E_{\mathrm{allbad}}],
\end{aligned}
\label{eq:process_reward}
\end{equation}
where all \(\rho\) coefficients are positive. This makes both useful Bad-to-Good progress and regression explicit at the trajectory level.

The process reward \(S_p\) scores transitions between evaluated candidates \(v_{1:m}\). It rewards recovery after an ineffective edit and penalizes a Good-to-Bad transition even if a later edit partially recovers. The terminal reward \(S_f^V\) scores the final candidate using its fresh rendering and the fixed reference. The interaction cost \(C_b\) penalizes redundant edits, rendering calls, and delayed stopping.

\begin{table}[H]
\centering
{\footnotesize
\setlength{\tabcolsep}{4pt}
\renewcommand{\arraystretch}{1.05}
\begin{tabularx}{\columnwidth}{@{}>{\raggedright\arraybackslash}p{0.19\columnwidth}>{\raggedright\arraybackslash}X@{}}
\toprule
\textbf{Reward term} & \textbf{Principal evidence and optimization effect} \\
\midrule
Good: \(S_k\) & Format validity and normalized equality of \(p_T\) and \(p_0\); rewards preservation. \\
\addlinespace[2pt]
Good: \(C_g\) & Edits, turns, and rendering calls; penalizes unnecessary changes and tool use. \\
\addlinespace[2pt]
Bad: \(S_f^V\) & Reference similarity, exact recovery, and the frozen Verifier; rewards terminal quality. \\
\addlinespace[2pt]
Bad: \(S_p\) & Recovery, stable-Good, regression, and all-Bad events in Eq.~\eqref{eq:process_reward}. \\
\addlinespace[2pt]
Bad: \(C_b\) & Repeated edits, render-free turns, render overuse, and budget exhaustion. \\
\bottomrule
\end{tabularx}}
\caption{Reward design for independent preservation and recovery objectives.}
\label{tab:reward_specification}
\end{table}

Contract-invalid completions receive an isolated validity penalty without renderer or Verifier calls, separating format shaping from semantic trajectory rewards.

\subsection{GRPO Objective}
For each prompt, GRPO samples a group of trajectories and normalizes their rewards:
\begin{equation}
\widehat A_i=
\frac{R_i-\overline R}
{\operatorname{std}(R_1,\ldots,R_G)+\epsilon}.
\label{eq:grpo_advantage}
\end{equation}
Let \(r_{i,t}(\theta)\) be the token-level likelihood ratio between the updated and rollout policies, and let \(\pi_{\mathrm{ref}}=\pi_{\theta_0}\) be the repair policy frozen at GRPO initialization. The objective combines the clipped group-relative advantage with a reference-policy KL constraint:
\begin{align}
\ell_{i,t}^{\mathrm{clip}}
=\min\!\big(&r_{i,t}\widehat A_i,\nonumber\\[-2pt]
&\operatorname{clip}(r_{i,t},1-\epsilon_c,1+\epsilon_c)\widehat A_i\big).
\label{eq:grpo_clip}
\end{align}
\begin{equation}
\mathcal L_{\mathrm{GRPO}}
=-\mathbb E\!\left[
\ell_{i,t}^{\mathrm{clip}}
-\beta D_{\mathrm{KL}}(\pi_\theta\,\|\,\pi_{\mathrm{ref}})
\right].
\label{eq:grpo_loss}
\end{equation}
The frozen reference policy is distinct from the frozen reward Verifier; Table~\ref{tab:optimization_config} reports the remaining settings.

\section{Additional Ablation Studies}
\label{app:pending_ablation_tables}

\subsection{Input-Modality Ablation}
Each input variant is trained separately for one epoch using the same Qwen3.5-9B initialization and data split. For \(I+R\), we deterministically align the predicted error context to the frozen OCR source and keep the predicted verdicts and types unchanged.

\begin{table}[H]
\centering
{
\scriptsize
\setlength{\tabcolsep}{1.5pt}
\begin{tabular*}{\columnwidth}{@{\extracolsep{\fill}}lcccccc@{}}
\toprule
\textbf{Input} & \textbf{Acc} & \textbf{Bad-F1} & \textbf{Type-F1} & \textbf{Loc-T} & \textbf{Loc-F} & \textbf{Eq-FP} \\
\midrule
\(I+p\) & 93.00 & 92.83 & 73.53 & \textbf{95.02} & \textbf{75.75} & 8.86 \\
\(I+R\) & 91.56 & 91.04 & 71.50 & 80.54 & 55.25 & 5.38 \\
\(I+p+R\) & \textbf{94.78} & \textbf{94.59} & \textbf{74.88} & 92.72 & 73.26 & \textbf{4.75} \\
\bottomrule
\end{tabular*}
}
\caption{Input-modality ablation. Eq-FP is the false-positive rate on rendering-equivalent positives.}
\label{tab:input_ablation}
\end{table}

The editable source most strongly supports exact localization, while rendering improves validity and equivalence robustness. Thus, \(I+p\) localizes best, but the complete triplet gives the strongest validity, type, and equivalence results.

\subsection{Iteration and Updated-Rendering Ablation}
All settings receive \((I,p_0,R(p_0))\), and Updated \(R\) denotes post-edit feedback. It adds 3.61/3.39 one-step ExactFix/VisFix points; iteration alone falls to 66.82\% ExactFix, while combining both reaches 82.17/86.23\% ExactFix/VisFix.

\begin{table}[H]
\vspace{-4pt}
\centering
{
\footnotesize
\setlength{\tabcolsep}{4pt}
\begin{tabular*}{\columnwidth}{@{\extracolsep{\fill}}ccccc@{}}
\toprule
\textbf{Iter.} & \textbf{Updated \(R\)} & \textbf{ExactFix} & \textbf{VisFix} & \textbf{Preserve} \\
\midrule
\(\times\) & \(\times\) & 72.46 & 77.20 & 91.47 \\
\(\times\) & \checkmark & 76.07 & 80.59 & 92.12 \\
\checkmark & \(\times\) & 66.82 & 74.94 & 90.81 \\
\checkmark & \checkmark & \textbf{82.17} & \textbf{86.23} & \textbf{93.44} \\
\bottomrule
\end{tabular*}
}
\caption{Iteration/rendering ablation.}
\label{tab:agent_structure_ablation}
\vspace{-4pt}
\end{table}

The combination of iteration and updated rendering yields the strongest repair and preservation scores. Relative to one-step repair without feedback, it raises ExactFix, VisFix, and Preserve by 9.71, 9.03, and 1.97 points. Iteration without updated rendering lowers those scores by 5.64, 2.26, and 0.66 points against the same baseline.

\clearpage
\section{Additional Evaluation Analyses}
\label{app:additional_evaluation}

\subsection{Diagnosis by Class and Preservation Outcomes}
Table~\ref{tab:class_and_preservation} reports sample-level diagnosis metrics for Good and all five error classes. The main-text Type-F1 is the macro-average over the three local error classes.

\begin{table}[H]
\centering
\caption{Diagnosis and preservation on OCRErrBench. Panel (a) reports percentages on the 900-case benchmark. In panel (b), Preserve and the edit audit cover the 457 Good inputs; VisFix is measured on Bad inputs.}
\label{tab:class_and_preservation}
{\scriptsize
\setlength{\tabcolsep}{2pt}
\begin{tabular*}{\columnwidth}{@{\extracolsep{\fill}}lrrr@{}}
\toprule
\multicolumn{4}{@{}l}{\textit{(a) DocEDR diagnosis by class}} \\
\textbf{Class} & \textbf{Precision} & \textbf{Recall} & \textbf{F1} \\
\midrule
Good & 93.25 & 96.72 & 94.95 \\
Invalid output & 70.97 & 73.33 & 72.13 \\
Global mismatch & 78.05 & 80.00 & 79.01 \\
Completeness & 79.51 & 77.60 & 78.54 \\
Content & 73.11 & 79.09 & 75.98 \\
Structure & 77.88 & 63.77 & 70.12 \\
\bottomrule
\end{tabular*}

\vspace{4pt}
\begin{tabular*}{\columnwidth}{@{\extracolsep{\fill}}lrrrr@{}}
\toprule
\multicolumn{5}{@{}l}{\textit{(b) Repair and preservation outcomes}} \\
\textbf{Method} & \shortstack[r]{\textbf{VisFix}\\\textbf{(\%)}} &
\shortstack[r]{\textbf{Preserve}\\\textbf{(\%)}} &
\shortstack[r]{\textbf{Benign}\\\textbf{edits}} &
\shortstack[r]{\textbf{Harmful}\\\textbf{edits}} \\
\midrule
Qwen3.5-397B & 72.23 & 94.75 & 19 & 5 (1.09\%) \\
DocEDR & 86.23 & 93.44 & 18 & 12 (2.63\%) \\
\bottomrule
\end{tabular*}}
\end{table}

DocEDR improves VisFix by 14.00 points over Qwen3.5-397B while retaining a similar Preserve score (93.44\% vs.\ 94.75\%). Of DocEDR's 12 harmful edits on Good inputs, 11 introduced rendering errors through changes to delimiters, escapes, or commands; one introduced an incorrect word boundary. A stricter initial threshold for routing inputs to repair could reduce harmful edits; the goal is to do so without losing the repair gains.

\newpage
\subsection{Refinement on External Document Text}
We evaluate the final GRPO-trained DocEDR policy on 4,760 OmniDocBench text blocks from 130 source documents excluded from OCRErrBench. The four parsers use the same region crops and reference texts. Because the regions are already localized, MinerU2.5 uses its second-stage text recognizer directly.

We apply identical text normalization before and after repair. Per-block normalized edit distance (NED) divides the Levenshtein distance by the longer of the normalized prediction and reference. Mean NED weights blocks equally, whereas character-weighted NED sums edit distances and divides by the sum of their length denominators. Both measures decrease for all four parsers, and only 0.06--0.48\% of blocks per parser have a higher NED after repair (Table~\ref{tab:omnidocbench_text_refinement}).

\begin{table}[H]
\centering
\caption{External text refinement on OmniDocBench. NED is on a \([0,1]\) scale. The changes in parentheses use unrounded values. The last percentage is the fraction of all 4,760 blocks with a higher post-repair NED.}
\label{tab:omnidocbench_text_refinement}
{\scriptsize
\setlength{\tabcolsep}{2pt}
\newcommand{\nedcell}[3]{\shortstack[r]{\(#1 \to \mathbf{#2}\)\\\((#3)\)}}
\newcommand{\outcomecell}[2]{\shortstack[r]{#1\\(#2\%)}}
\renewcommand{\arraystretch}{1.23}
\begin{tabular*}{\columnwidth}{@{\extracolsep{\fill}}lrrr@{}}
\toprule
\textbf{Parser} &
\shortstack[r]{\textbf{NED} $\downarrow$\\Before $\to$ After} &
\shortstack[r]{\textbf{Char-weighted}\\\textbf{NED} $\downarrow$\\Before $\to$ After} &
\shortstack[r]{\textbf{Improved /}\\\textbf{Unchanged /}\\\textbf{Degraded}} \\
\midrule
MinerU2.5 & \nedcell{0.0093}{0.0091}{-0.0001} & \nedcell{0.0180}{0.0179}{-0.0002} & \outcomecell{67 / 4,676 / 17}{0.36} \\[3pt]
PaddleOCR-VL-1.5 & \nedcell{0.0085}{0.0085}{-0.00002} & \nedcell{0.0077}{0.0077}{-0.00002} & \outcomecell{21 / 4,736 / 3}{0.06} \\[3pt]
InternVL3.5-8B & \nedcell{0.0130}{0.0106}{-0.0024} & \nedcell{0.0221}{0.0186}{-0.0036} & \outcomecell{220 / 4,517 / 23}{0.48} \\[3pt]
GLM-4.6V-Flash & \nedcell{0.0120}{0.0103}{-0.0017} & \nedcell{0.0342}{0.0322}{-0.0020} & \outcomecell{356 / 4,388 / 16}{0.34} \\
\bottomrule
\end{tabular*}}
\end{table}

\begingroup
\renewcommand{\thefigure}{B\arabic{figure}}
\setcounter{figure}{0}

\begin{figure*}[p]
\centering
\includegraphics[width=0.94\textwidth]{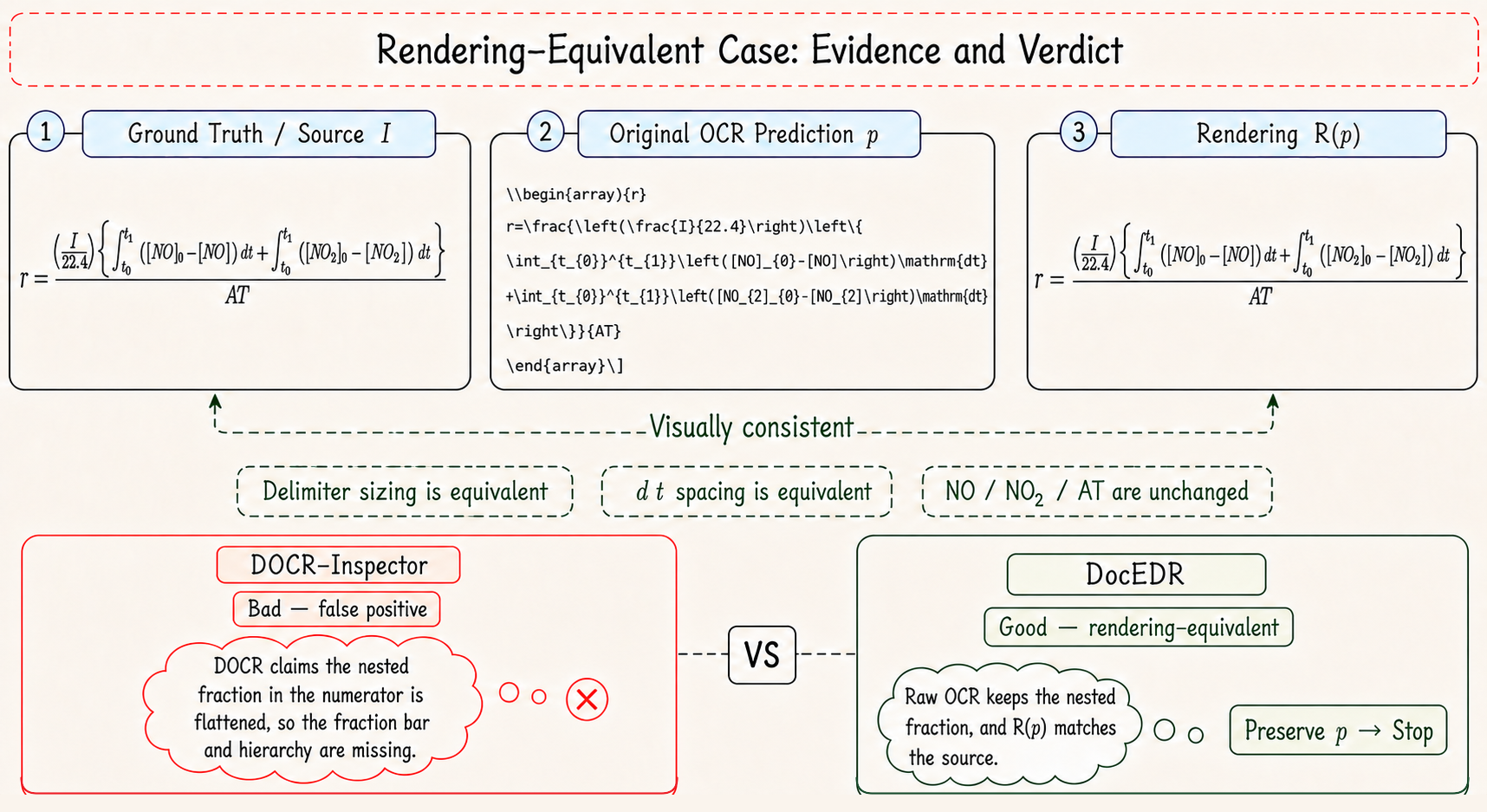}
\caption{Rendering-equivalent formula case with alternative delimiter sizing. DocEDR preserves the valid OCR source, whereas DOCR-Inspector incorrectly reports a missing fraction structure.}
\label{fig:equivalent_case_1}
\end{figure*}

\begin{figure*}[p]
\centering
\includegraphics[width=0.94\textwidth]{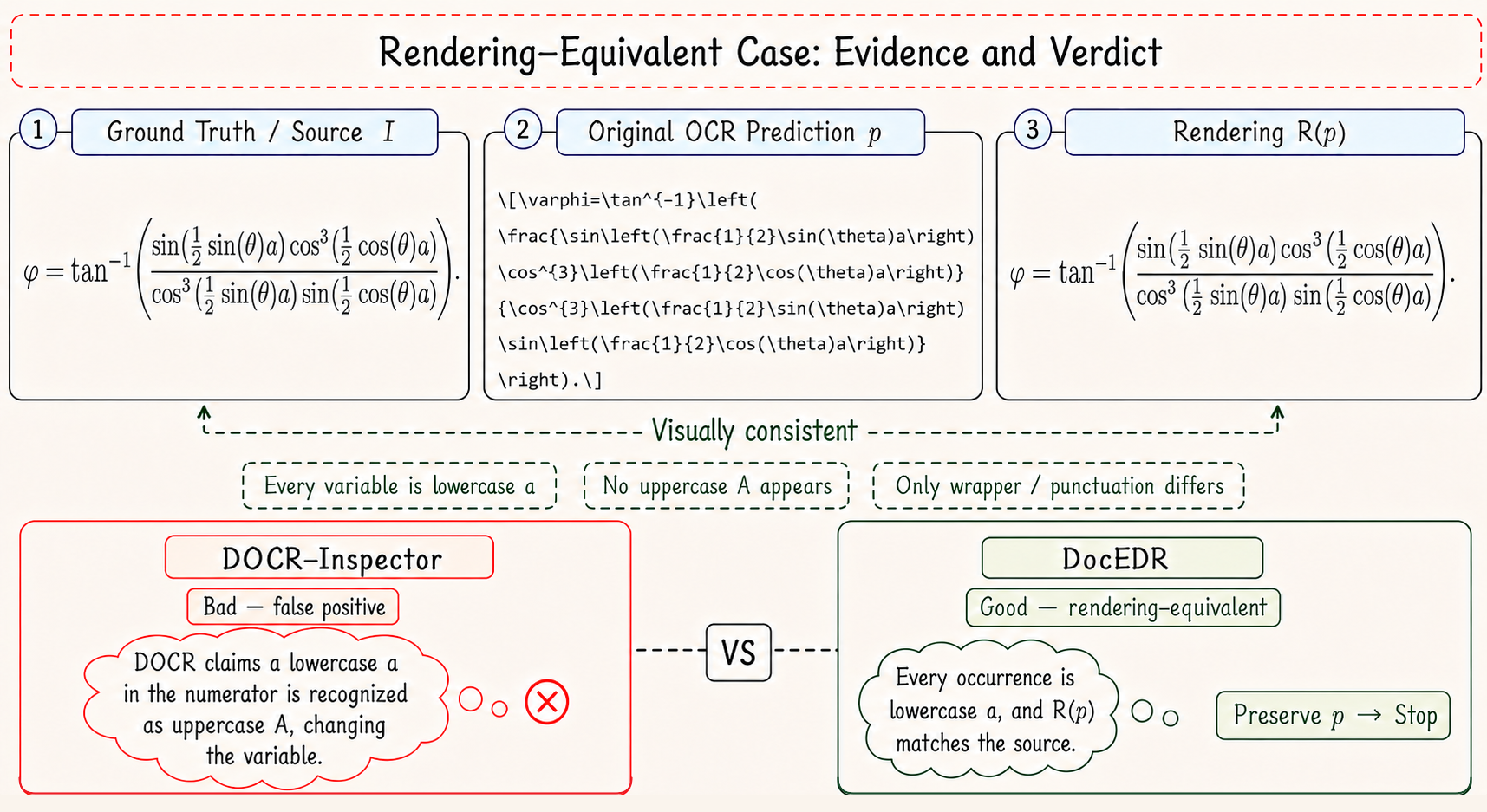}
\caption{Rendering-equivalent formula case with wrapper and punctuation variation. DocEDR preserves the valid OCR source, whereas DOCR-Inspector hallucinates a variable change.}
\label{fig:equivalent_case_2}
\end{figure*}

\begin{figure*}[p]
\centering
\includegraphics[width=0.94\textwidth]{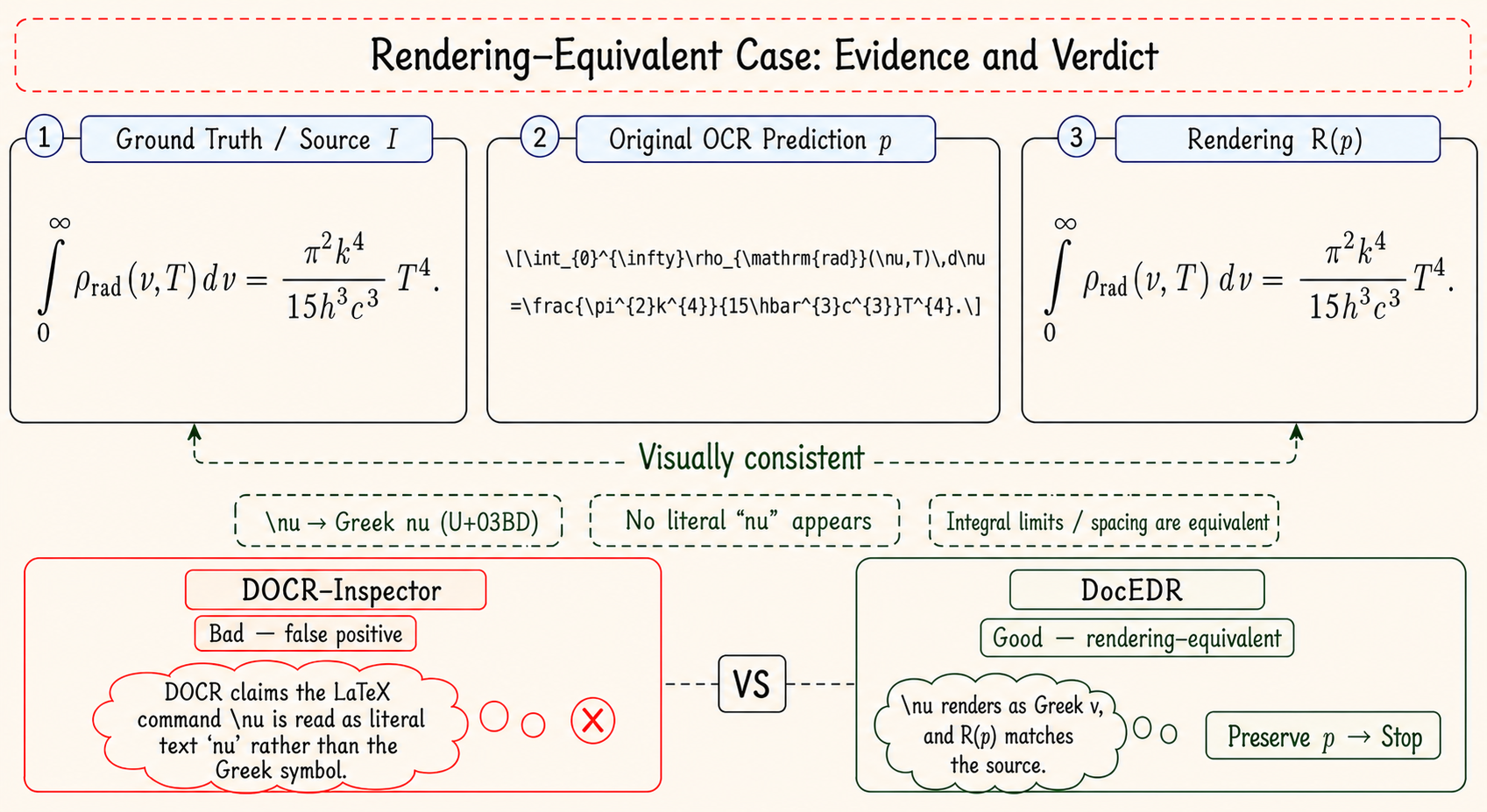}
\caption{Rendering-equivalent formula case using a \LaTeX{} command for the Greek symbol \(\nu\). DocEDR resolves the command through rendering, whereas DOCR-Inspector incorrectly treats it as literal text.}
\label{fig:equivalent_case_3}
\end{figure*}

\begin{figure*}[p]
\centering
\includegraphics[width=0.94\textwidth]{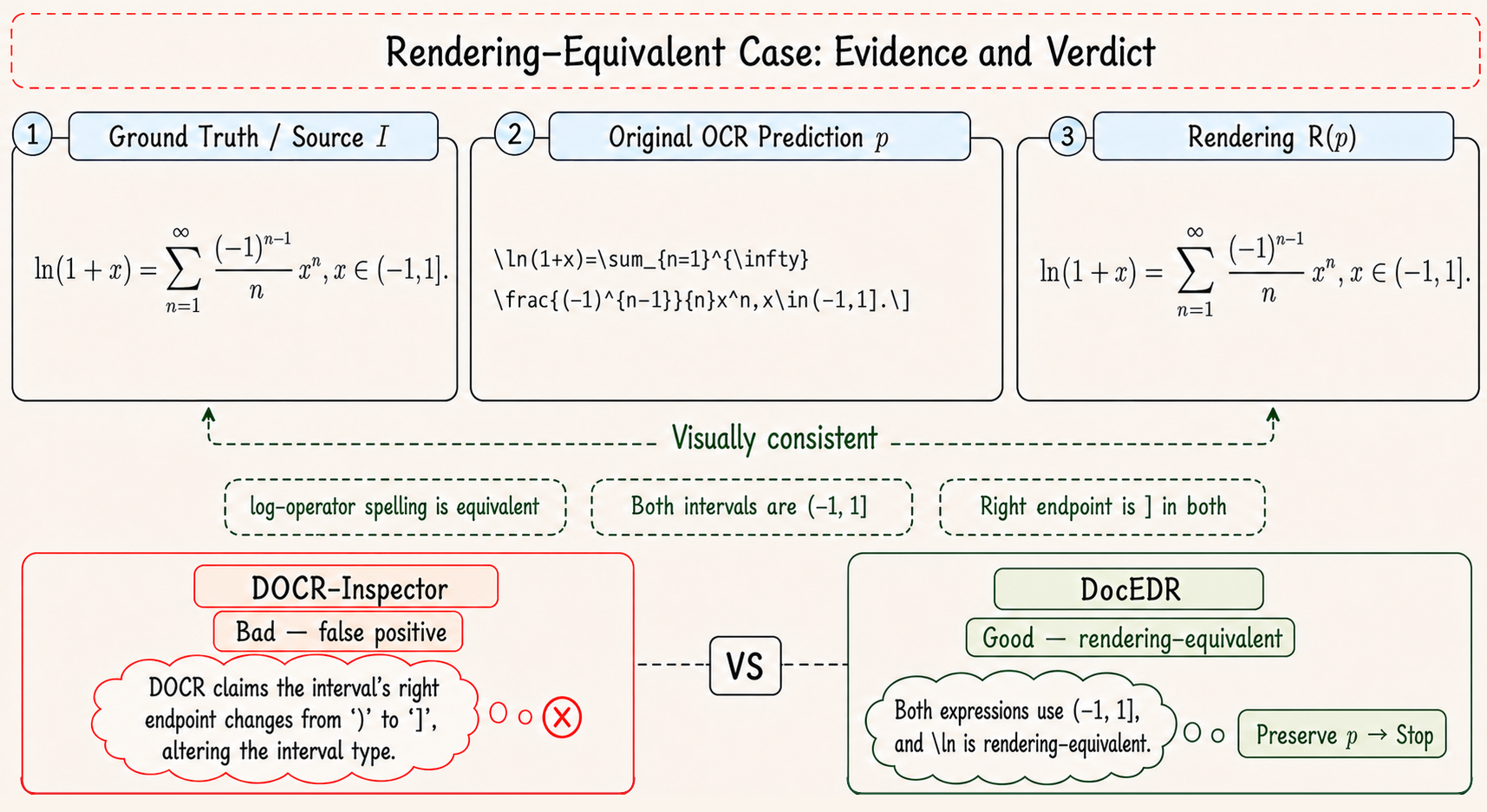}
\caption{Rendering-equivalent formula case with operator spelling and interval-wrapper variation. DocEDR preserves the valid OCR source, whereas DOCR-Inspector incorrectly reports a changed endpoint.}
\label{fig:equivalent_case_4}
\end{figure*}

\begin{figure*}[p]
\centering
\includegraphics[width=0.94\textwidth]{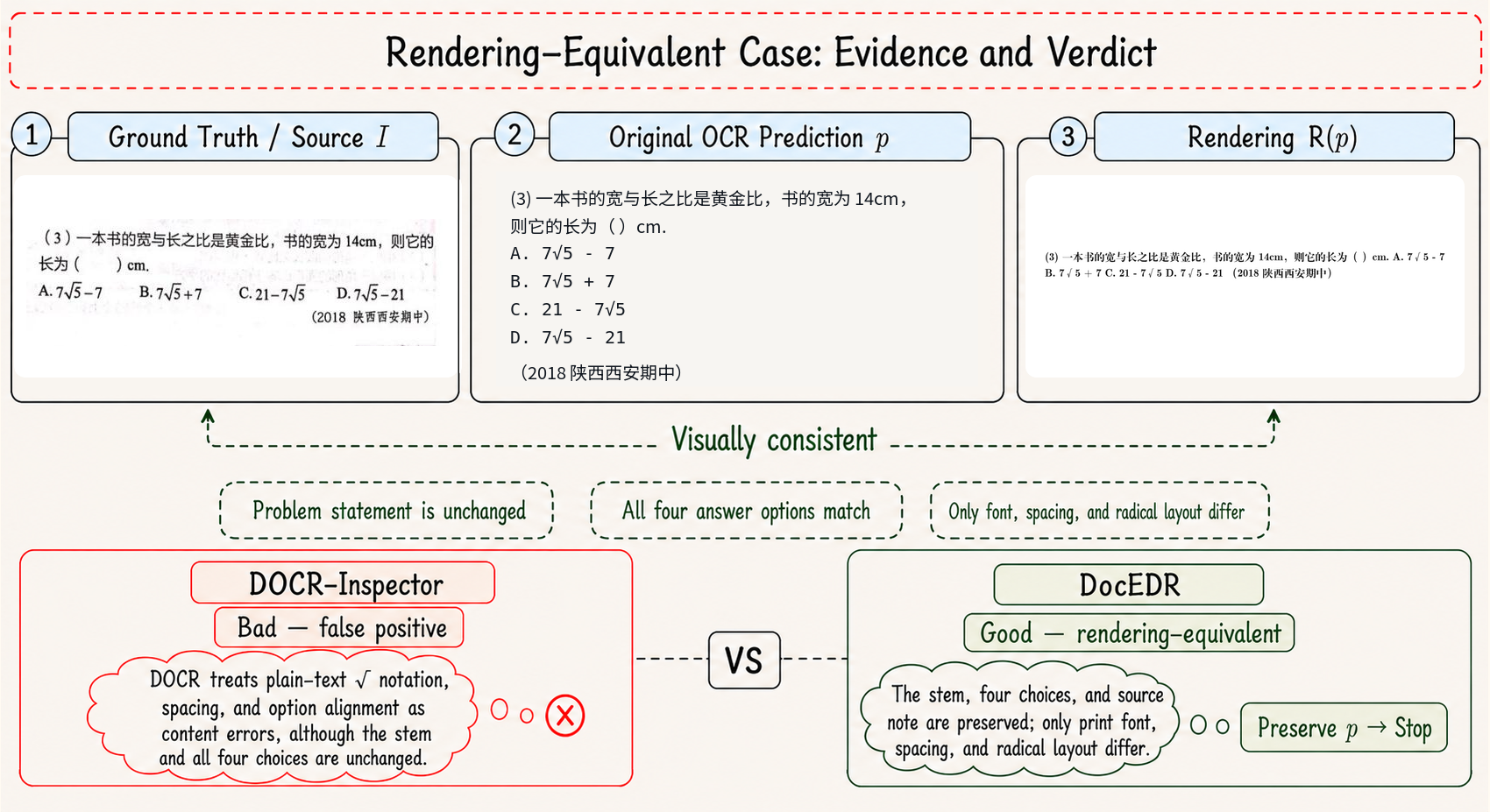}
\caption{Rendering-equivalent Chinese text case. The problem statement, four answer choices, and source note are preserved despite benign font, spacing, and radical-layout differences; DocEDR therefore stops without editing.}
\label{fig:equivalent_case_5}
\end{figure*}

\begin{figure*}[p]
\centering
\includegraphics[width=0.94\textwidth]{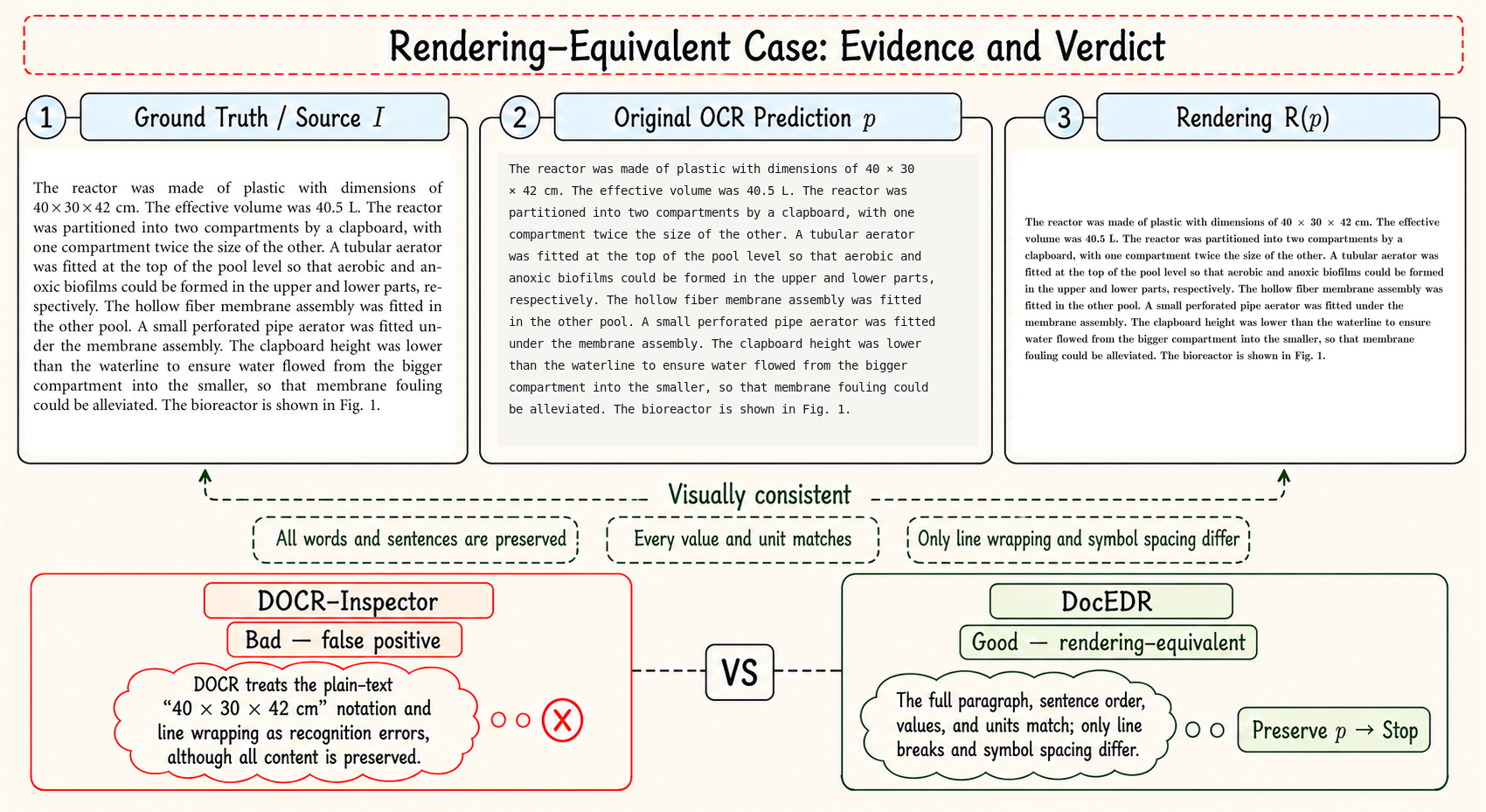}
\caption{Rendering-equivalent English paragraph case. All words, values, units, and sentence order are preserved despite benign line-wrapping and symbol-spacing differences; DocEDR therefore stops without editing.}
\label{fig:equivalent_case_6}
\end{figure*}

\endgroup

\end{document}